\documentclass[11pt]{article}

\usepackage{fontspec}
\defaultfontfeatures{Ligatures=TeX}

\usepackage[margin=1in]{geometry}
\usepackage{amsmath}
\usepackage{amssymb}

\usepackage{graphicx}
\graphicspath{{figures/}}
\usepackage{booktabs}
\usepackage{array}
\usepackage{multirow}
\usepackage{longtable}
\usepackage{caption}
\usepackage{xcolor}
\usepackage{float}

\usepackage{tikz}
\usetikzlibrary{arrows.meta,positioning,shapes.geometric,fit,backgrounds}

\usepackage{authblk}

\usepackage[numbers,sort&compress]{natbib}

\usepackage{microtype}

\usepackage[colorlinks=true,linkcolor=blue!50!black,citecolor=blue!50!black,urlcolor=blue!50!black]{hyperref}

\title{\textbf{Generative Augmentation for EEG Motor Imagery Classification: A Class-Conditional VAE with Cycle-Consistent Decoder Refinement}}

\author[1]{Matei Moldoveanu}
\author[1]{Alain Sirois}
\author[1]{Claire Ben Ali}
\author[2]{Fabien Lotte}
\author[3]{Florian Yger}

\affil[1]{Naqi Logix, Paris, France}
\affil[2]{Inria Center at Univ. Bordeaux / LaBRI, Talence, France}
\affil[3]{LITIS, INSA Rouen-Normandy, France}

\date{}

\begin{document}

\maketitle
\vspace{-2em}
\begin{center}
\small E-mail: \href{mailto:matei@naqilogix.com}{matei@naqilogix.com} and \href{mailto:alain@naqilogix.com}{alain@naqilogix.com}
\end{center}
\vspace{1em}

\begin{abstract}
We investigate whether a generative model can supply useful synthetic motor-imagery (MI) electroencephalography (EEG) trials that improve the accuracy of independent downstream classifiers. We train a class-conditional variational autoencoder (CVAE) with an integrated latent classifier on the Zhou motor-imagery dataset, using the learned per-class prior as a generator: sampling the prior for a given label and decoding it into a synthetic, label-consistent signal. A constraint on the covariance matrix of the generated data encourages preservation of covariance structure, and the model is trained with a schedule that alternates ordinary VAE training with a decoder-focused phase that sharpens the generative pathway used for augmentation.

We measure the effect of adding synthetic trials to the training set under two evaluation protocols --- within-user (pooled 60/20/20 split across subjects) and cross-user (leave-one-subject-out, LOSO) --- across four representative EEG classification pipelines: Common Spatial Patterns with Linear Discriminant Analysis (CSP+LDA), tangent-space features with a Support Vector Machine (TGSP+SVM), Minimum Distance to Riemannian Mean (MDM), and a neural network based on EEGNetv4 (henceforth EEGNet).

Results are aggregated across independent augmentation draws, random seeds (within-user), or leave-one-subject-out folds (cross-user), with uncertainty reported as 95\% confidence intervals (Student's $t$-distribution) computed over per-seed/per-fold averages.

We find that synthetic EEG from the CVAE is most credible as a source of class-structured, covariance-like data rather than as a substitute for real raw EEG: it can raise the point estimate for MDM, but the broader augmentation claim remains conservative --- observed gains are small and classifier-dependent.

\end{abstract}

\section{Introduction}

EEG-based brain--computer interfaces (EEG-BCIs) provide a non-invasive way to read neural activity and translate it into control signals. Among BCI paradigms, MI is a widely studied task and the focus of this work. A major challenge in MI-BCI research is that EEG data collection is time-consuming, and signals exhibit high variability across sessions and subjects. This variability makes training robust classifiers difficult. One promising mitigation is data augmentation, which generates additional training samples from existing EEG recordings.

Recent advances in generative neural networks have enabled increasingly realistic EEG data synthesis. Variational autoencoders (VAEs) are attractive for this purpose because they provide an explicit, samplable latent distribution rather than only an implicit generator, which makes it possible to condition generation on class label and to inspect the structure of what the model has learned. This work investigates whether a class-conditional VAE~\citep{sohn2015learning} can generate motor-imagery EEG trials that measurably improve the accuracy of downstream classifiers.

Prior work has generally kept two design choices separate: VAEs have been trained for EEG synthesis without an explicit covariance constraint, and covariance-aware constraints have instead been explored mainly alongside GAN-based generators, so a class-conditional VAE combined with an explicit covariance objective remains largely unexplored. A different line of work sidesteps this combination by moving the generative model itself onto the covariance manifold: the Riemannian Geometry-Preserving VAE (RGP-VAE) trains directly on EEG spatial-covariance matrices, using Riemannian-aware mappings to keep generated covariances valid symmetric positive-definite matrices~\citep{polaka2026riemannian}. The present work takes a third route: the VAE continues to operate on raw time-series EEG in ordinary Euclidean space, while a covariance constraint is applied to the decoded output, so covariance structure is encouraged without requiring the VAE itself to leave Euclidean space.

This paper makes four main contributions. First, we introduce a class-conditional VAE architecture for five-channel motor-imagery EEG that jointly optimises for a reconstruction objective, a Log-Euclidean spatial-covariance constraint and a trainable per-class Gaussian prior usable directly as a sampler. The training procedures use a lightweight cycle-consistency refinement of the decoder's generative pathway. Second, we empirically evaluate the resulting synthetic data as a training-set augmentation across four downstream classifiers (CSP+LDA, TGSP+SVM, MDM, and EEGNet) and two generalisation regimes (within-user and cross-user/LOSO). Third, we present a data-quality analysis, spanning latent-space, signal-time, signal-frequency, and spatial-covariance perspectives, that separates the question of whether the generator learns class structure from whether that structure helps a downstream classifier. Finally, we report seven single-factor ablation studies isolating the contribution of the training schedule, reconstruction loss, class-separation mechanism, covariance constraint, and their relative loss weightings.

Section~\ref{sec:related} surveys related work on generative EEG augmentation, covariance-aware and Riemannian-manifold generative models, and disentangled latent representations. Section~\ref{sec:method} then describes the model architecture, training objective, and training procedure. Section~\ref{sec:setup} describes the evaluation protocols and downstream classifiers. Section~\ref{sec:results} reports results on augmentation impact and generated-data quality. Section~\ref{sec:discussion} discusses the findings, and Section~\ref{sec:limitations} states the study's limitations. Section~\ref{sec:conclusion} concludes. Appendix~\ref{app:ablation} reports ablation studies and Appendix~\ref{app:persubject} reports per-seed/per-subject results underlying the aggregated tables in Section~\ref{sec:results}.

\section{Related Work}
\label{sec:related}

Two broad families of generative neural networks have been used for EEG data synthesis. Variational autoencoders (VAEs) encode data into a latent space and reconstruct it via a decoder; once trained, new samples can be generated by manipulating the latent codes. VAEs are effective at capturing global data structure, but they sometimes produce overly smooth signals~\citep{bethge2022eeg2vec,zancanaro2024veegnet,cisotto2024hveegnet,zhao2024vaeeg}. Generative adversarial networks (GANs), by contrast, use a generator--discriminator setup that produces sharper outputs but often requires large datasets for training~\citep{doku2026structure,williams2025eeggan}. Hybrid approaches leverage the strengths of both: for example, Trans-cVAE-GAN~\citep{yao2025transcvaegan} combines a VAE encoder for EEG data with a GAN that generates samples from the latent space, producing high-fidelity EEG while preserving meaningful latent structure. Purpose-built software toolkits such as EEG-GAN aim to make this GAN-based augmentation approach directly usable for improving downstream neural classification.

A parallel strand of work makes the generative model itself Riemannian-aware, rather than treating covariance information only as an auxiliary loss on Euclidean output. The Riemannian Geometry-Preserving VAE (RGP-VAE)~\citep{polaka2026riemannian} trains an encoder--decoder pair whose inputs and outputs are EEG spatial-covariance matrices rather than raw time series: it combines a Riemannian-distance reconstruction term, a tangent-space reconstruction term, and a diversity term, and relies on geometric mappings such as parallel transport to keep generated covariances valid symmetric positive-definite matrices and to learn a subject-invariant latent space for MI-BCI augmentation.

The present work sits at the intersection of these threads: it trains a latent classifier jointly with a class-conditional VAE, and it explicitly targets the reconstruction's channel covariance rather than reconstruction fidelity alone. Doku et al.~\citep{doku2026structure} pursue a structurally similar goal with an adversarial rather than variational generator, embedding a Riemannian-conditional GAN so that both the generator and discriminator respect the SPD-manifold structure of EEG covariance during structure-preserving augmentation.

\section{Method}
\label{sec:method}

\subsection{Design Rationale}

The model jointly optimises three complementary objectives. The first, reconstruction, aims to learn a faithful generative model of EEG and combines two terms: a Soft-DTW (dynamic time warping) loss~\citep{cuturi2017softdtw}, which encourages accurate reconstruction of the EEG waveform in the time domain and is preferred over MSE because it is robust to the small temporal shifts and local temporal distortions common in EEG signals, and a covariance constraint, which preserves the covariance structure of the reconstructed signals, keeping them physiologically plausible and suitable for covariance-based classifiers operating in Riemannian space.

The second objective, latent class separation, aims to learn well-defined, class-specific latent regions that can be reliably sampled. It combines a KL divergence constraint, which regularises the latent representations by encouraging samples from the same class to form compact, well-structured distributions while remaining close to their class prior, with a latent classifier, which encourages separation between classes in the latent space and thereby enables a learnable class-conditioned prior.

The third objective, cycle consistency, trains the decoder to generate signals consistent with the encoder's learned representation. It comprises a latent MSE term, which ensures that a signal generated from a sampled latent vector is mapped back by the encoder to the same latent location and so promotes latent-space consistency, and a latent classification term, which ensures that reconstructed signals are encoded back into the same class from which their latent representation was sampled, preserving class identity during generation.

The cycle-consistency objective aims to address two limitations of standard VAE training. First, during training the decoder reconstructs latent codes produced by the encoder, whereas augmentation generates samples by decoding latent vectors drawn directly from the learned class-conditional prior. Since these latent distributions are not identical, the decoder may not be optimally trained for the latent vectors encountered during generation. A decoder-focused phase therefore trains the decoder directly on prior samples, improving the quality of generated data, particularly in low-data settings where augmentation is most valuable. Second, the cycle-consistency loss encourages generated signals to encode back to the original latent representation and class, improving agreement between the encoder and decoder. However, optimising only this objective risks over-specialising the decoder to the encoder; alternating between joint optimisation and decoder-focused training balances these competing objectives. Appendix~\ref{app:schedule} evaluates whether this strategy provides a measurable benefit.

This refinement is related to prior VAE decoder-refinement work such as those of Jha et al.~\citep{jha2018disentangling}, Silvestri and Ambrogioni~\citep{silvestri2025covae}, and Caillon and Esling~\citep{caillon2021rave}. RAVE~\citep{caillon2021rave} is closest to our decoder-focused phase --- its second stage likewise freezes the encoder and trains only the decoder --- but it does so on posterior samples from real audio via a genuinely adversarial objective in a one-time, non-repeating switch, whereas ours operates on prior samples, in a non-adversarial way (just a latent MSE and classification-consistency loss), and alternates back into joint training every two epochs (Appendix~\ref{app:schedule}).

\subsection{Model Architecture}

The model is a CVAE with an integrated latent classifier. Real trials are encoded to a Gaussian posterior, sampled, and decoded; the same latent space is read by a small classifier; and a learned per-class prior provides both the KL target during training and the sampling distribution for generation.

\begin{figure}[htbp]
\centering
\begin{tikzpicture}[
  node distance=3mm and 5mm,
  box/.style={draw, rounded corners, align=center, minimum width=20mm, minimum height=5mm, font=\small, thick, draw=indigo},
  >=Latex
]
\definecolor{indigo}{RGB}{67,56,202}

\node[box] (x) {EEG trial\\$x \in \mathbb{R}^{1\times5\times500}$};
\node[box, below=of x] (enc) {Encoder\\temporal + spatial conv};
\node[box, below left=3mm and 5mm of enc] (mu) {$\mu \in \mathbb{R}^{100}$};
\node[box, below right=3mm and 5mm of enc] (logvar) {$\log\sigma^2 \in \mathbb{R}^{100}$};
\node[box, below=16mm of enc] (reparam) {Reparameterise\\$z=\mu+\sigma\odot\varepsilon,\ \varepsilon\sim\mathcal{N}(0,I)$};
\node[box, below=of reparam] (z) {latent $z \in \mathbb{R}^{100}$};
\node[box, below left=9mm and 4mm of z] (dec) {Decoder\\(conv upsample)};
\node[box, below right=9mm and 4mm of z] (cls) {Latent classifier};
\node[box, below=of dec] (xhat) {reconstruction $\hat{x}\in\mathbb{R}^{1\times5\times500}$};
\node[box, below=of cls] (yhat) {$\hat{y}$ (3 classes)};

\draw[->] (x) -- (enc);
\draw[->] (enc) -- (mu);
\draw[->] (enc) -- (logvar);
\draw[->] (mu) -- (reparam);
\draw[->] (logvar) -- (reparam);
\draw[->] (reparam) -- (z);
\draw[->] (z) -- (dec);
\draw[->] (z) -- (cls);
\draw[->] (dec) -- (xhat);
\draw[->] (cls) -- (yhat);
\end{tikzpicture}
\caption{Model architecture. The encoder maps a trial to a Gaussian posterior; a latent sample is decoded and classified. The class-conditional prior is trainable per class.}
\label{fig:architecture}
\end{figure}
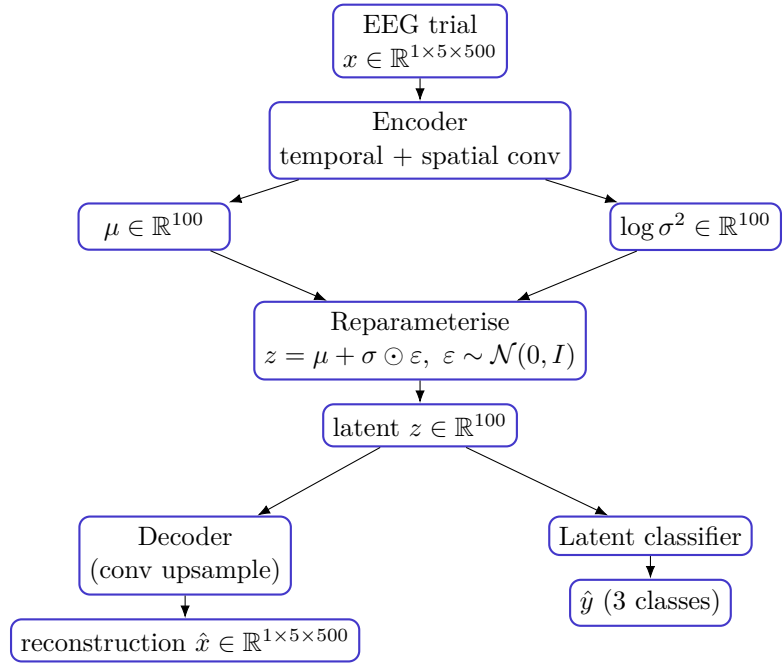

Trials are 3-class (feet / left-hand / right-hand), 5 channels, 500 time samples (5\,s at 100\,Hz), so the input tensor is $1 \times 5 \times 500$. The latent space is 100-dimensional.

\subsubsection{Encoder}

The encoder maps an input trial to posterior mean and log-variance vectors.

\begin{table}[H]
\centering
\caption{Encoder (input $1 \times 5 \times 500$).}
\label{tab:encoder}
\small
\begin{tabular}{@{}lp{8.7cm}c@{}}
\toprule
Stage & Operation & Output C$\times$H$\times$W \\
\midrule
Input & --- & 1 $\times$ 5 $\times$ 500 \\
Temporal conv & Conv2d 1$\to$8, kernel (1, 50); BatchNorm & 8 $\times$ 5 $\times$ 451 \\
Spatial conv & Depthwise Conv2d 8$\to$16, kernel (5, 1); BN; ELU; AvgPool (1, 2); Dropout 0.5 & 16 $\times$ 1 $\times$ 225 \\
Grouped block 1 & Grouped Conv2d 16$\to$16, kernel (5, 50), same-pad; BN; ELU; AvgPool (1, 2); Dropout & 16 $\times$ 1 $\times$ 112 \\
Grouped block 2 & Grouped Conv2d 16$\to$16, kernel (5, 50), same-pad; BN; ELU; Dropout & 16 $\times$ 1 $\times$ 112 \\
Separable conv & Conv2d 16$\to$16, kernel (1, 1); BN; ELU; Dropout & 16 $\times$ 1 $\times$ 112 \\
Flatten & --- & 1792 \\
Posterior heads & Linear 1792$\to$100 ($\mu$) and Linear 1792$\to$100 ($\log\sigma^2$) & 100, 100 \\
\bottomrule
\end{tabular}
\end{table}

The temporal convolution uses valid padding. The two grouped convolutions and the separable convolution use \texttt{padding="same"}.

\subsubsection{Decoder}

The decoder maps a latent vector to a reconstructed trial.

\begin{table}[H]
\centering
\caption{Decoder (input $z\in \mathbb{R}^{100}$).}
\label{tab:decoder}
\small
\begin{tabular}{@{}llc@{}}
\toprule
Stage & Operation & Output C$\times$H$\times$W \\
\midrule
Project & Linear 100$\to$3968; reshape & 32 $\times$ 2 $\times$ 62 \\
Block 1 & Upsample $\times$(2, 2); Conv2d 32$\to$32, kernel (3, 7); BN; ELU & 32 $\times$ 4 $\times$ 124 \\
Block 2 & Upsample $\to$ (5, 500); Conv2d 32$\to$16, kernel (5, 9); BN; ELU & 16 $\times$ 5 $\times$ 500 \\
Block 3 & Conv2d 16$\to$8, kernel (3, 7); BN; ELU & 8 $\times$ 5 $\times$ 500 \\
Output & Conv2d 8$\to$1, kernel (1, 1) & 1 $\times$ 5 $\times$ 500 \\
\bottomrule
\end{tabular}
\end{table}

Both \texttt{Upsample} stages use bilinear interpolation (\texttt{align\_corners=False}). The three convolutions in Blocks 1--3 use \texttt{padding="same"}. The output layer has no activation function.

\subsubsection{Class-Conditional Prior}

Each class has a trainable mean and log-variance vector in latent space ($\mu_p(y), \sigma_p^2(y) \in \mathbb{R}^{100}$). For generation, $z \sim \mathcal{N}(\mu_p(y),\sigma_p^2(y))$ is decoded into a synthetic trial labelled $y$.

\subsubsection{Latent Classifier}

\begin{table}[H]
\centering
\caption{Latent classifier (input $z\in \mathbb{R}^{100}$).}
\label{tab:latent-classifier}
\small
\begin{tabular}{@{}llc@{}}
\toprule
Stage & Operation & Output \\
\midrule
Input & Latent sample & 100 \\
Hidden 1 & Linear 100$\to$64; ReLU; Dropout 0.5 & 64 \\
Hidden 2 & Linear 64$\to$64; ReLU; Dropout 0.5 & 64 \\
Hidden 3 & Linear 64$\to$64; ReLU; Dropout 0.5 & 64 \\
Output & Linear 64$\to$3 & 3 logits \\
\bottomrule
\end{tabular}
\end{table}

\subsection{Training Objective}

The model is trained to minimise a weighted sum of four terms.

\medskip
\noindent\textbf{Reconstruction --- Soft-DTW.}

\begin{equation}
\mathcal{L}_{\text{rec}} = \operatorname{softDTW}_\gamma(\hat{x}, x), \qquad \gamma = 1
\label{eq:rec}
\end{equation}

\noindent\textbf{Latent regularisation --- conditional KL.}

\begin{equation}
\mathcal{L}_{\text{KL}} = \tfrac{1}{2}\,\mathbb{E}\!\left[\log\sigma_p^2 - \log\sigma^2 + \frac{\sigma^2 + (\mu - \mu_p)^2}{\sigma_p^2} - 1\right]
\label{eq:kl}
\end{equation}

\noindent\textbf{Spatial-covariance match --- Log-Euclidean.}

\begin{equation}
\mathcal{L}_{\text{cov}} = \big\lVert \log \Sigma(\hat{x}) - \log \Sigma(x) \big\rVert_F^2
\label{eq:cov}
\end{equation}

Here $\Sigma(\cdot)$ is the sample channel covariance ($5 \times 5$) of a trial --- de-meaned over the time axis, normalised by $T-1$, with no shrinkage (no OAS/Ledoit--Wolf) --- regularised to be strictly SPD by adding $\epsilon I$ with $\epsilon = 10^{-6}$. $\log(\cdot)$ is the matrix logarithm via eigendecomposition; this is the log-Euclidean distance between SPD matrices in their tangent-space (log) representation. This penalty is inspired by the structure-preserving loss of Doku et al.'s Riemannian conditional GAN~\citep{doku2026structure}, which enforces covariance-matrix fidelity on the generator's output.

\medskip
\noindent\textbf{Classification --- cross-entropy.}

\begin{equation}
\mathcal{L}_{\text{cls}} = -\sum_{c} \mathbf{1}[y = c]\,\log \operatorname{softmax}(g(z))_c
\label{eq:cls}
\end{equation}

\noindent\textbf{Total objective:}

\begin{equation}
\mathcal{L} = 0.01\,\mathcal{L}_{\text{rec}} + 5.0\,\mathcal{L}_{\text{KL}} + 1.5\,\mathcal{L}_{\text{cov}} + 1.0\,\mathcal{L}_{\text{cls}}
\label{eq:total}
\end{equation}

\subsection{Training Procedure}

Optimisation uses Adam, learning rate $10^{-3}$, batch size 30. The model is trained under an \textbf{alternating schedule} that switches between two phases in 2-epoch blocks.

\paragraph{Standard phase.} All model components (encoder, decoder, class-conditional prior, and classifier) are optimised jointly using the full training objective.

\paragraph{Decoder-focused phase.} The encoder, prior, and classifier are kept fixed (no gradient updates to their weights) and used only to compute the consistency loss below; only the decoder is updated. For each class, latent samples are drawn from the corresponding class prior, decoded into synthetic EEG trials, and passed back through the (fixed) encoder. The decoder is optimised to (i) minimise the squared error between the original latent sample and this re-encoded representation and (ii) preserve the class label when the (fixed) latent classifier reads the re-encoded representation. This prior$\to$decode$\to$re-encode cycle encourages consistency between the learned prior and the generated signals used for augmentation.

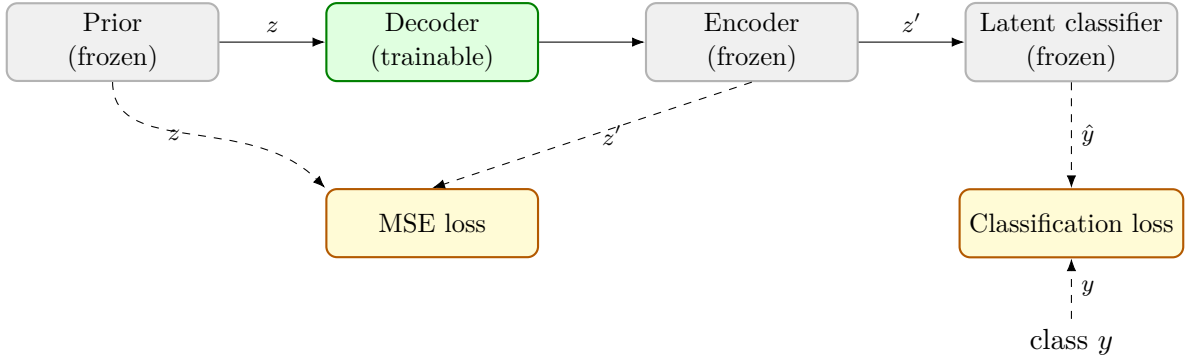
\begin{figure}[htbp]
\centering
\begin{tikzpicture}[
  node distance=10mm and 14mm,
  box/.style={draw, rounded corners, align=center, minimum width=28mm, minimum height=9mm, font=\small, thick},
  frozen/.style={box, fill=gray!12, draw=gray!60},
  trainable/.style={box, fill=green!12, draw=green!50!black},
  loss/.style={box, fill=yellow!20, draw=orange!70!black},
  >=Latex
]
\node[frozen] (prior) {Prior\\(frozen)};
\node[trainable, right=of prior] (dec) {Decoder\\(trainable)};
\node[frozen, right=of dec] (enc) {Encoder\\(frozen)};
\node[frozen, right=of enc] (clf) {Latent classifier\\(frozen)};

\node[loss, below=14mm of dec] (mse) {MSE loss};
\node[loss, below=14mm of clf] (ce) {Classification loss};
\node[below=8mm of ce] (y) {class $y$};

\draw[->] (prior) -- node[above,font=\footnotesize]{$z$} (dec);
\draw[->] (dec) -- (enc);
\draw[->] (enc) -- node[above,font=\footnotesize]{$z'$} (clf);

\draw[->,dashed] (prior.south) .. controls +(0,-10mm) and +(-10mm,10mm) .. node[left,font=\footnotesize]{$z$} (mse.north west);
\draw[->,dashed] (enc.south) -- node[right,font=\footnotesize]{$z'$} (mse.north);
\draw[->,dashed] (clf.south) -- node[right,font=\footnotesize]{$\hat{y}$} (ce.north);
\draw[->,dashed] (y) -- node[right,font=\footnotesize]{$y$} (ce.south);
\end{tikzpicture}
\caption{Decoder-focused phase, prior$\to$decode$\to$re-encode cycle. Encoder, prior, and classifier are frozen (grey); only the decoder (green) receives gradient updates.}
\label{fig:decoder-focused}
\end{figure}

In the matched 3-seed ablation reported in Appendix~\ref{app:schedule}, the alternating schedule was associated with improved point estimates for the usefulness of generated data for augmentation: it moved MDM's with-Aug delta from net-negative to net-positive and raised the accuracy of classifiers trained solely on synthetic data.

\section{Experimental Setup}
\label{sec:setup}

\subsection{Dataset and Evaluation Protocols}

Two evaluation protocols are used, both on the same Zhou2016 dataset~\citep{zhou2016fully} (3 classes: feet/left-hand/right-hand; 5 sensorimotor channels C3, Cz, C4, FCz, CPz; bandpass 8--32\,Hz; resampled to 100\,Hz). Inputs are z-scored using training-set statistics only --- no leakage into validation or test.

\paragraph{Within-user evaluation.} Trials are partitioned \textbf{60/20/20} into train/validation/test \emph{within each (subject, session) group}, so every group contributes to all three splits while no trial is shared. The real training pool for downstream classifiers is \textbf{train + validation}. The split is controlled by a random seed, treated as an experimental variable in the multi-seed study.

\paragraph{Cross-user evaluation (Leave-One-Subject-Out, LOSO).} For each fold, \textbf{one subject is withheld as the test set; the other three subjects form the train+val set, split 75/25 within each (subject, session) pair}. Standardisation uses training-set statistics only --- applied to the held-out subject without ever observing it (the realistic deployment condition). Representative fold sizes (held-out subject 1): \textbf{987 train / 334 val / 479 test}. All four subjects are held out in turn (4 folds).

\subsection{Downstream Classifiers}

For each evaluation protocol, four standard EEG pipelines are trained independently of the VAE and scored on the respective held-out test set:

\begin{table}[H]
\centering
\caption{Downstream classification pipelines.}
\label{tab:pipelines}
\small
\begin{tabular}{@{}ll@{}}
\toprule
Name & Pipeline \\
\midrule
CSP+LDA & Common Spatial Patterns (8) $\to$ LDA \\
TGSP+SVM & Covariances $\to$ Tangent space (Riemann) $\to$ SVM (RBF) \\
MDM & Covariances $\to$ Minimum-Distance-to-Mean (Riemann) \\
EEGNet & EEGNet v4 (deep convolutional network) \\
\bottomrule
\end{tabular}
\end{table}

\paragraph{EEGNet architecture and training.} EEGNet is based on braindecode's EEGNetv4, wrapped in an sklearn-compatible estimator.

\begin{table}[H]
\centering
\caption{EEGNet architecture, as configured for this evaluation.}
\label{tab:eegnet}
\small
\begin{tabular}{@{}p{2.6cm}p{4.0cm}p{7.2cm}@{}}
\toprule
Block & Layer & Config \\
\midrule
Input & --- & \texttt{(batch, 5 channels, 500 time samples)} at 100\,Hz this revision \\
1.Temporal conv & Conv2d, 1 $\to$ F1 & kernel $(1, 200)$, \texttt{same} padding, no bias, BatchNorm \\
2.Spatial (depthwise) conv & Depthwise Conv2d, F1 $\to$ F1$\times$D, groups=F1 & kernel $(5, 1)$ (one tap per channel), max-norm constraint = 1, BatchNorm $\to$ ELU $\to$ AvgPool $(1, 4)$ $\to$ Dropout \\
3.Separable conv & Depthwise Conv2d (kernel $(1, 16)$) + pointwise Conv2d, F1$\times$D $\to$ F2 & BatchNorm $\to$ ELU $\to$ AvgPool $(1, 8)$ $\to$ Dropout \\
4.Classifier & Conv2d, F2 $\to$ 3 classes & \texttt{final\_conv\_length="auto"} (kernel spans the remaining time axis) \\
\bottomrule
\end{tabular}
\end{table}

\textbf{Config used:} F1 = 6, D = 3, F2 = 6, temporal kernel length = 200 samples (2\,s at 100\,Hz), depthwise kernel = 16, pool sizes 4/8, dropout = 0.25, ELU activation, spatial-conv max-norm = 1 (all library defaults except F1/D/F2/kernel\_length, which are set explicitly).

\textbf{Training setup:} Adam optimiser, learning rate $10^{-2}$, batch size 30, cross-entropy loss, up to 50 epochs. Model selection uses a checkpoint callback that saves the model with the best accuracy on an internal validation split (50\% of the training data provided by the condition) and reloads that checkpoint after training completes.

\subsection{Augmentation Protocol}

For each classifier, three training conditions are compared on the held-out test set: a baseline condition using real training data only, a with-Aug condition using real data plus synthetic trials drawn from the class-conditional prior, and an Aug-only condition using synthetic trials alone, which probes whether the generator carries class structure on its own.

The augmentation set contains \textbf{the same number of samples as the original training set} (1$\times$ volume) for the main results. Each augmentation condition is repeated over 4 independent draws; results are reported as mean $\pm$ 95\% CI. (For the within-user multi-seed study, the underlying run recorded 5 draws per seed; only the first 4 are used, matching the cross-user LOSO study's draw count.)

\emph{Note: classifiers are trained and tested on the same data as the CVAE, i.e.\ band-passed, resampled, and standardised on the training dataset; the synthetic data is used as generated by the model.}

\section{Results}
\label{sec:results}

\subsection{Summary of Findings}

Synthetic EEG from the CVAE model is most credible as a source of class-structured, covariance-like data --- not as a replacement for real raw EEG. It can help MDM in point estimate, but the main augmentation claim should remain conservative: the observed gains are small and classifier-dependent.

\textbf{Accuracy.} For most classification algorithms, augmentation yields no improvement beyond the observed variability and, in some cases, reduces performance (full results in Table~\ref{tab:t5}, within-user, and Table~\ref{tab:t8}, cross-user). MDM is the only classifier for which augmentation can produce a measurable benefit: in some within-user seeds, MDM performance increases with augmentation, and training on augmented-only data can even outperform training on the combined real-and-augmented dataset, though these gains are not consistently reproduced across all random seeds. A similar pattern is observed in the cross-user evaluation, where some held-out users benefit from augmentation while others show little or no improvement.

\begin{figure}[htbp]
\centering
\includegraphics[width=0.85\textwidth]{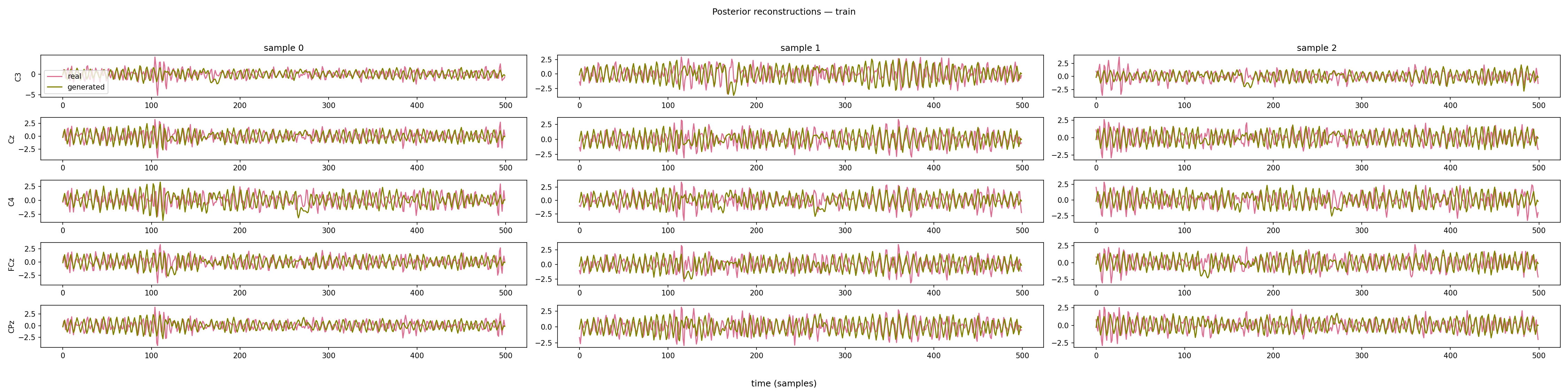}\\[0.5em]
\includegraphics[width=0.85\textwidth]{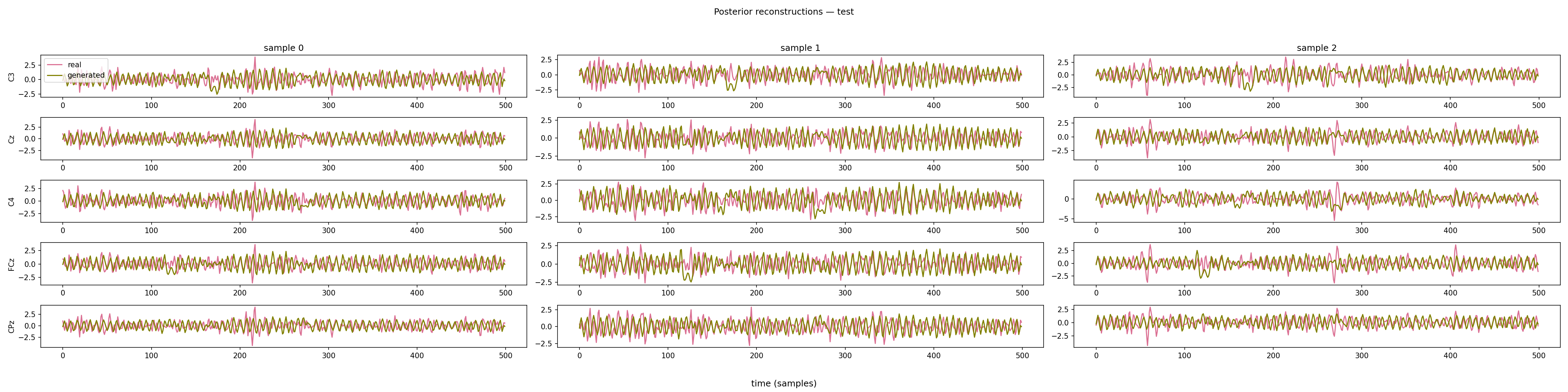}
\caption{Posterior reconstructions (top: train trials; bottom: test trials; real in pink, reconstruction in orange; rows are channels).}
\label{fig:reconstructions}
\end{figure}

\textbf{Data quality.} Two data-quality takeaways follow. First, the learned latent space captures meaningful class structure: the learned class priors confirm that this structure generalises to unseen data, as classifying held-out test trials by the nearest learned prior mean achieves approximately 0.8 accuracy (within-user), in line with the other classification algorithms, and the t-SNE visualisation further shows samples generated from the learned class priors occupying the same broad class regions as real encoded samples (Section~\ref{sec:latent-quality}). Second, signal reconstructions capture some of the mu-band power but are poor otherwise: posterior reconstructions have a relative RMSE of around 1.42 on both train and test, whereas generated feet-class spectra place their dominant peak in the 8--15\,Hz mu band (Figures~\ref{fig:reconstructions}, \ref{fig:fft-single}, and~\ref{fig:fft-trial-avg}); spectral amplitude is not faithful, with generated mu-band peaks for the feet class roughly twice as tall as real peaks and full-spectrum correlations of 0.42--0.47 across channels; spatial covariance is partially recovered, with the generated per-channel variance pattern for the feet class correlating with real train data at 0.91 against a train-vs-test correlation of 0.97; and covariance-based classifiers reach over 0.7 accuracy when trained on synthetic data only, whereas EEGNet performs much worse, though still above chance, under the same condition.

\subsection{Impact of Data Augmentation}

This section examines the impact of adding generated trials to a training set on downstream classifier accuracy, for the four classifiers introduced in Section~\ref{sec:setup}. The augmentation set in the main tables contains the \textbf{same number of samples as the original training set} (1$\times$ volume).

\textbf{How to read the tables.} The \textbf{Train} column reports the test-set accuracy (mean $\pm$ 95\% CI) obtained by training on the real training data only, where the CI reflects variability across seeds (within-user) or held-out subjects (cross-user). The \textbf{with-Aug} and \textbf{Aug-only} columns report the change in test-set accuracy relative to the Train baseline ($\Delta$ = condition $-$ Train mean) when training on the corresponding dataset, together with the 95\% CI of the delta.

\subsubsection{Within-User Results}

\textbf{Multi-seed analysis (4 augmentation draws; seeds 0, 1, 2).} To account for variability arising from the random train/test split and model initialisation, the CVAE is trained independently for three random seeds (0, 1, and 2). Each seed produces a different data partition and a different trained model. For each trained model, the augmentation analysis is repeated over four independent augmentation draws, where a new synthetic dataset is generated and each downstream classifier is retrained from scratch. The reported results are the mean and 95\% confidence interval computed across the three per-seed means.

\begin{table}[H]
\centering
\caption{Synthetic data impact, within-user, cross-seed.}
\label{tab:t5}
\small
\begin{tabular}{@{}lccc@{}}
\toprule
Classifier & Baseline (mean $\pm$ 95\% CI) & $\Delta$(with-Aug) $\pm$ 95\% CI & $\Delta$(Aug-only) $\pm$ 95\% CI \\
\midrule
CSP+LDA & 0.776 $\pm$ 0.028 & +0.004 $\pm$ 0.038 & $-$0.035 $\pm$ 0.074 \\
TGSP+SVM & 0.812 $\pm$ 0.017 & $-$0.004 $\pm$ 0.020 & $-$0.089 $\pm$ 0.017 \\
MDM & 0.648 $\pm$ 0.008 & +0.014 $\pm$ 0.024 & +0.045 $\pm$ 0.093 \\
EEGNet & 0.832 $\pm$ 0.037 & +0.010 $\pm$ 0.046 & $-$0.350 $\pm$ 0.147 \\
\bottomrule
\end{tabular}
\end{table}

Across the three independent runs, none of the with-Aug deltas have a 95\% confidence interval excluding zero, indicating that any change brought by the synthetic data is small relative to the variability introduced by different train/test splits and model initialisations.

Since Table~\ref{tab:t5} suggests that adding one train-sized batch of synthetic data provides, at most, a modest benefit, the next question is whether increasing the amount of synthetic data changes this behaviour. Table~\ref{tab:t6} therefore evaluates augmentation volumes ranging from \textbf{0.5$\times$} to \textbf{5$\times$} the size of the original training set.

\begin{table}[H]
\centering
\caption{Within-user test-accuracy delta vs. no augmentation, by volume.}
\label{tab:t6}
\small
\begin{tabular}{@{}lcccc@{}}
\toprule
Volume ($\times$ train) & CSP+LDA & TGSP+SVM & MDM & EEGNet \\
\midrule
0 (Baseline) & 0.776 $\pm$ 0.028 & 0.812 $\pm$ 0.017 & 0.648 $\pm$ 0.008 & 0.832 $\pm$ 0.037 \\
0.5 ($\Delta$ vs. 0$\times$) & $-$0.001 $\pm$ 0.024 & $-$0.004 $\pm$ 0.024 & +0.002 $\pm$ 0.021 & +0.004 $\pm$ 0.021 \\
1 ($\Delta$ vs. 0$\times$) & +0.002 $\pm$ 0.046 & +0.004 $\pm$ 0.028 & +0.029 $\pm$ 0.039 & $-$0.001 $\pm$ 0.032 \\
2 ($\Delta$ vs. 0$\times$) & $-$0.001 $\pm$ 0.070 & $-$0.003 $\pm$ 0.037 & +0.028 $\pm$ 0.043 & $-$0.009 $\pm$ 0.066 \\
5 ($\Delta$ vs. 0$\times$) & +0.005 $\pm$ 0.052 & $-$0.007 $\pm$ 0.014 & +0.038 $\pm$ 0.064 & $-$0.009 $\pm$ 0.021 \\
\bottomrule
\end{tabular}
\end{table}

None of the augmentation volumes produce a statistically significant improvement over the no-augmentation baseline. However, the point estimates reveal a consistent trend for MDM: performance increases from essentially no change at \textbf{0.5$\times$} (+0.002) to approximately \textbf{+3--4 percentage points} for augmentation volumes of \textbf{1$\times$} and above, after which the gains plateau. In contrast, CSP+LDA, TGSP+SVM, and EEGNet remain centred close to zero across all augmentation volumes, indicating little sensitivity to the amount of synthetic data. Although the confidence intervals are too wide to confirm the MDM trend statistically, the consistency of the positive point estimates across all augmentation volumes suggests that MDM is the classifier most likely to benefit from additional synthetic training data.

\textbf{Latent-space classifiers.} Classifiers are also trained directly on the encoder's posterior mean and are evaluated alongside the signal-space pipelines above, using the same three seed-0/1/2 models as Table~\ref{tab:t5} (1$\times$ augmentation volume, 4 draws per seed, real base = train+val).

\begin{table}[H]
\centering
\caption{Latent-space classifiers, within-user, cross-seed.}
\label{tab:t7}
\small
\begin{tabular}{@{}lccc@{}}
\toprule
Classifier (on $\mu$) & Baseline (mean $\pm$ 95\% CI) & $\Delta$(with Aug) $\pm$ 95\% CI & $\Delta$(Aug only) $\pm$ 95\% CI \\
\midrule
LDA & 0.833 $\pm$ 0.078 & +0.002 $\pm$ 0.035 & $-$0.009 $\pm$ 0.035 \\
SVM-RBF & 0.826 $\pm$ 0.026 & +0.019 $\pm$ 0.034 & $-$0.014 $\pm$ 0.030 \\
Logistic Regression & 0.841 $\pm$ 0.059 & +0.007 $\pm$ 0.039 & $-$0.018 $\pm$ 0.038 \\
Nearest prior mean & 0.814 $\pm$ 0.048 & --- & --- \\
\bottomrule
\end{tabular}
\end{table}

The latent-space results mirror the signal-space findings: with-Aug produces only small improvements whose confidence intervals include zero, indicating no significant benefit across the three seeds. However, unlike the signal-space pipelines, classifiers trained directly on the encoder's latent representation remain close to their Train performance even when trained exclusively on synthetic samples, with Aug-only degradations below two percentage points. Together with the strong baseline performance of the latent classifiers, this suggests that the encoder learns a discriminative latent representation.

\subsubsection{Cross-User Results}

To measure how well the generator transfers to a subject it has never seen, the VAE is trained independently four times --- once per held-out subject --- each time withholding that subject's data entirely and training only on the other three. Each held-out subject therefore has its own trained model and its own test set. For each of these four trained models, the augmentation analysis is repeated over four independent augmentation draws, where a new synthetic dataset is generated and each downstream classifier is retrained from scratch.

\begin{table}[H]
\centering
\caption{Augmentation impact, cross-user, aggregated.}
\label{tab:t8}
\small
\begin{tabular}{@{}lccc@{}}
\toprule
Classifier & Baseline (mean $\pm$ 95\% CI) & $\Delta$(with Aug) $\pm$ 95\% CI & $\Delta$(Aug only) $\pm$ 95\% CI \\
\midrule
CSP+LDA & 0.725 $\pm$ 0.051 & +0.012 $\pm$ 0.032 & $-$0.033 $\pm$ 0.055 \\
TGSP+SVM & 0.724 $\pm$ 0.103 & $\approx$0 $\pm$ 0.015 & $-$0.107 $\pm$ 0.179 \\
MDM & 0.617 $\pm$ 0.082 & +0.009 $\pm$ 0.053 & +0.013 $\pm$ 0.121 \\
EEGNet & 0.741 $\pm$ 0.117 & $\approx$0 $\pm$ 0.023 & $-$0.226 $\pm$ 0.077 \\
\bottomrule
\end{tabular}
\end{table}

Across the four held-out subjects, none of the with-Aug deltas indicate an improvement from augmentation that is significant relative to the variability introduced by which subject is held out. Even the improvement seen for MDM in the within-user case disappears in the cross-user case.

Since Table~\ref{tab:t8} suggests that adding one train-sized batch of synthetic data provides, at most, a modest benefit, the next question is whether increasing the amount of synthetic data changes this behaviour. Table~\ref{tab:t9} therefore evaluates augmentation volumes ranging from \textbf{0.5$\times$} to \textbf{5$\times$} the size of the original training set.

\begin{table}[H]
\centering
\caption{Cross-user test-accuracy delta vs. no augmentation.}
\label{tab:t9}
\small
\begin{tabular}{@{}lcccc@{}}
\toprule
Volume ($\times$ train) & CSP+LDA & TGSP+SVM & MDM & EEGNet \\
\midrule
0 (Baseline) & 0.725 $\pm$ 0.051 & 0.724 $\pm$ 0.103 & 0.617 $\pm$ 0.082 & 0.741 $\pm$ 0.117 \\
0.5 ($\Delta$ vs. 0$\times$) & +0.003 $\pm$ 0.039 & $-$0.002 $\pm$ 0.026 & +0.008 $\pm$ 0.033 & +0.016 $\pm$ 0.098 \\
1 ($\Delta$ vs. 0$\times$) & +0.015 $\pm$ 0.041 & $-$0.003 $\pm$ 0.011 & +0.005 $\pm$ 0.072 & +0.018 $\pm$ 0.054 \\
2 ($\Delta$ vs. 0$\times$) & $-$0.002 $\pm$ 0.038 & $-$0.007 $\pm$ 0.011 & +0.004 $\pm$ 0.097 & +0.014 $\pm$ 0.059 \\
5 ($\Delta$ vs. 0$\times$) & $-$0.006 $\pm$ 0.040 & $-$0.004 $\pm$ 0.021 & +0.013 $\pm$ 0.110 & +0.009 $\pm$ 0.052 \\
\bottomrule
\end{tabular}
\end{table}

This shows that augmentation volume neither reliably improves nor harms the classification algorithms in the cross-user setting, across the whole range tested.

\textbf{Latent-space classifiers.} Classifiers trained directly on the encoder's posterior mean $\mu$ are also evaluated alongside the signal-space pipelines above, using the same four per-subject models as Table~\ref{tab:t8} (1$\times$ augmentation volume, 4 draws per fold), to assess whether the encoder learns user-agnostic features that generalise to unseen users.

\begin{table}[H]
\centering
\caption{Latent-space classifiers, cross-user.}
\label{tab:t10}
\small
\begin{tabular}{@{}lccc@{}}
\toprule
Classifier (on $\mu$) & Baseline (mean $\pm$ 95\% CI) & $\Delta$(with Aug) $\pm$ 95\% CI & $\Delta$(Aug only) $\pm$ 95\% CI \\
\midrule
LDA & 0.714 $\pm$ 0.096 & +0.037 $\pm$ 0.048 & +0.020 $\pm$ 0.047 \\
SVM-RBF & 0.711 $\pm$ 0.065 & +0.043 $\pm$ 0.047 & +0.027 $\pm$ 0.028 \\
Logistic Regression & 0.755 $\pm$ 0.060 & +0.005 $\pm$ 0.055 & $-$0.023 $\pm$ 0.062 \\
Nearest prior mean & 0.737 $\pm$ 0.114 & --- & --- \\
\bottomrule
\end{tabular}
\end{table}

The latent-space results broadly mirror the signal-space findings. Latent-space classification is able to extract class-specific information that is not user-dependent, at least to the same level as the covariance-based information.

\subsection{Data Quality of the Generated Samples}
\label{sec:latent-quality}

This section examines within-user generated-sample fidelity for the seed-0 model.

\subsubsection{Latent Space Representation}

The latent geometry is visualised using t-SNE~\citep{vandermaaten2008visualizing}, a dimensionality-reduction technique, to assess how the encoder organises trials in latent space and whether generated samples land in the same regions as real trials of the same class. This technique projects the 100-dimensional latent vectors down to 2 dimensions, so some distortion of the true separation is expected. A more robust way to assess latent separation is latent classifier accuracy, which was shown above (Tables~\ref{tab:t7} and~\ref{tab:t10}) to perform similarly to signal-space classifiers.

\begin{figure}[htbp]
\centering
\includegraphics[width=0.75\textwidth]{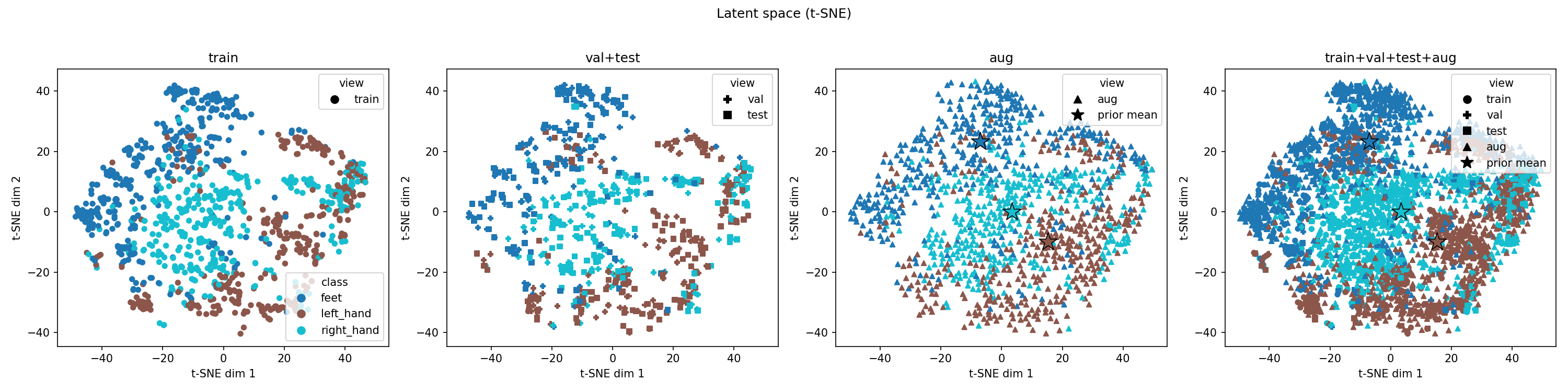}
\caption{Latent space (t-SNE). Latent posteriors for train, validation and test trials, the directly-sampled class-conditional prior (``aug''), and the per-class prior means (stars).}
\label{fig:tsne}
\end{figure}

The prior samples fall inside the same regions as real trials of each class --- a visual interpretation of the results in Tables~\ref{tab:t7} and~\ref{tab:t10}, where distance to the class prior mean alone is sufficient to reach a classification accuracy that matches the downstream classifiers. The same seed-0 model is examined more closely in Table~\ref{tab:t11}.

\begin{table}[H]
\centering
\caption{Latent classifiers vs.\ signal-based classifiers, within-user (seed 0). To show concretely how the latent representation compares with the signal-space classifiers, this table fits a latent-space classifier (SVM-RBF) and the unfitted nearest-prior-mean rule alongside the covariance-based signal-space classifiers (CSP+LDA, TGSP+SVM) on the same real training data, with and without 1$\times$ prior augmentation added, for this within-user model, scored on the same held-out test set.}
\label{tab:t11}
\small
\begin{tabular}{@{}llcc@{}}
\toprule
Classifier & Space & Train accuracy & with-Aug accuracy \\
\midrule
CSP+LDA & signal (covariance) & 0.769 & 0.785 \\
TGSP+SVM & signal (covariance) & 0.811 & 0.808 \\
SVM-RBF & latent ($\mu$) & 0.814 & 0.829 \\
Nearest prior mean & latent ($\mu$, unfitted) & 0.792 & --- \\
\bottomrule
\end{tabular}
\end{table}

Similarly to the signal-based classifiers, the latent-space ones are also largely unaffected by the prior data, showcasing that the augmentation adds little new information to the classification process for any of these classifiers at seed 0. The unfitted nearest-prior-mean rule has no with-Aug variant (it involves no training data or fitting at all, so there is nothing for augmentation to add to).

\subsubsection{Signal Time Representation}

The reconstruction fidelity test asks whether the decoder reconstructs the specific trial it was given --- as opposed to only producing plausible samples from the class prior, which is what augmentation actually uses. If reconstruction is not sufficiently good, the augmented data is unlikely to be useful. Comparing train and test reconstructions also indicates whether this capability generalises beyond the trials seen during training.

The decoder preserves the oscillatory pattern and envelope of each trial, but the reconstruction is a poor match to the specific real trial. RMSE and pointwise correlation are both alignment-sensitive metrics --- they penalise any timing shift as if it were a content error, which is exactly the failure mode Soft-DTW is designed to tolerate. To evaluate fidelity under the same metric the model is actually trained with, 20 real feet-class trials are drawn from the seed-0 model's train split, each is posterior-reconstructed (encode to $\mu$, decode), and the normalised Soft-DTW distance is computed between all 40 real and reconstructed trials, then visualised with a t-SNE embedding of that $40\times40$ distance matrix.

\begin{figure}[htbp]
\centering
\includegraphics[width=0.7\textwidth]{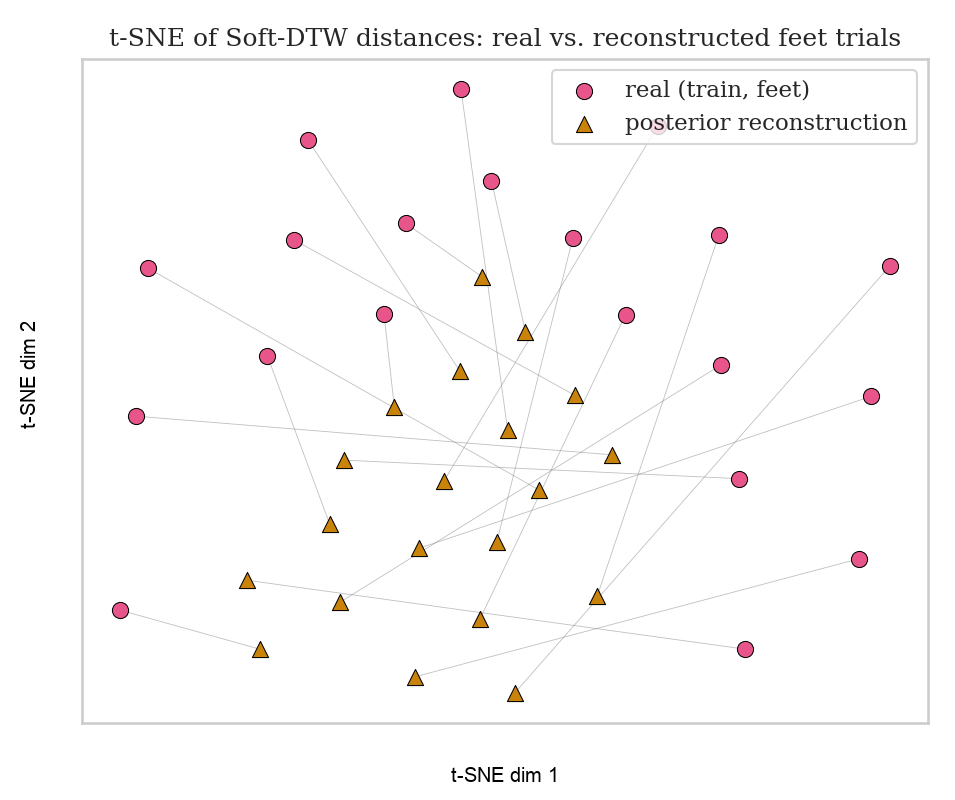}
\caption{t-SNE embedding of pairwise Soft-DTW distance, 20 real feet trials vs.\ their 20 posterior reconstructions (train split, seed 0; circles = real, triangles = reconstructions; grey lines connect each real trial to its own reconstruction). Distances are the normalised Soft-DTW metric the model is trained with, not Euclidean distance on the raw signal.}
\label{fig:dtw-tsne}
\end{figure}

\begin{table}[H]
\centering
\caption{Average pairwise Soft-DTW distance, 20 trials of feet class vs.\ their reconstructions.}
\label{tab:t12}
\small
\begin{tabular}{@{}lc@{}}
\toprule
Comparison & Mean Soft-DTW distance \\
\midrule
Real vs.\ real (20 trials, off-diagonal pairs) & 1883 \\
Reconstruction vs.\ reconstruction (20 trials, off-diagonal pairs) & 283 \\
Real vs.\ its own reconstruction (paired, 20 trials) & 1492 \\
\bottomrule
\end{tabular}
\end{table}

The reconstructions are \textbf{more tightly clustered among themselves} than the real trials are among each other --- visible directly in Figure~\ref{fig:dtw-tsne}. The real-vs-reconstruction distance sits closer to the real-vs-real scale than to the reconstruction-vs-reconstruction scale, meaning a reconstruction is typically about as Soft-DTW-distant from its own source trial as two arbitrary real trials are from each other.

\subsubsection{Signal Frequency Representation}

\begin{figure}[htbp]
\centering
\includegraphics[width=0.85\textwidth]{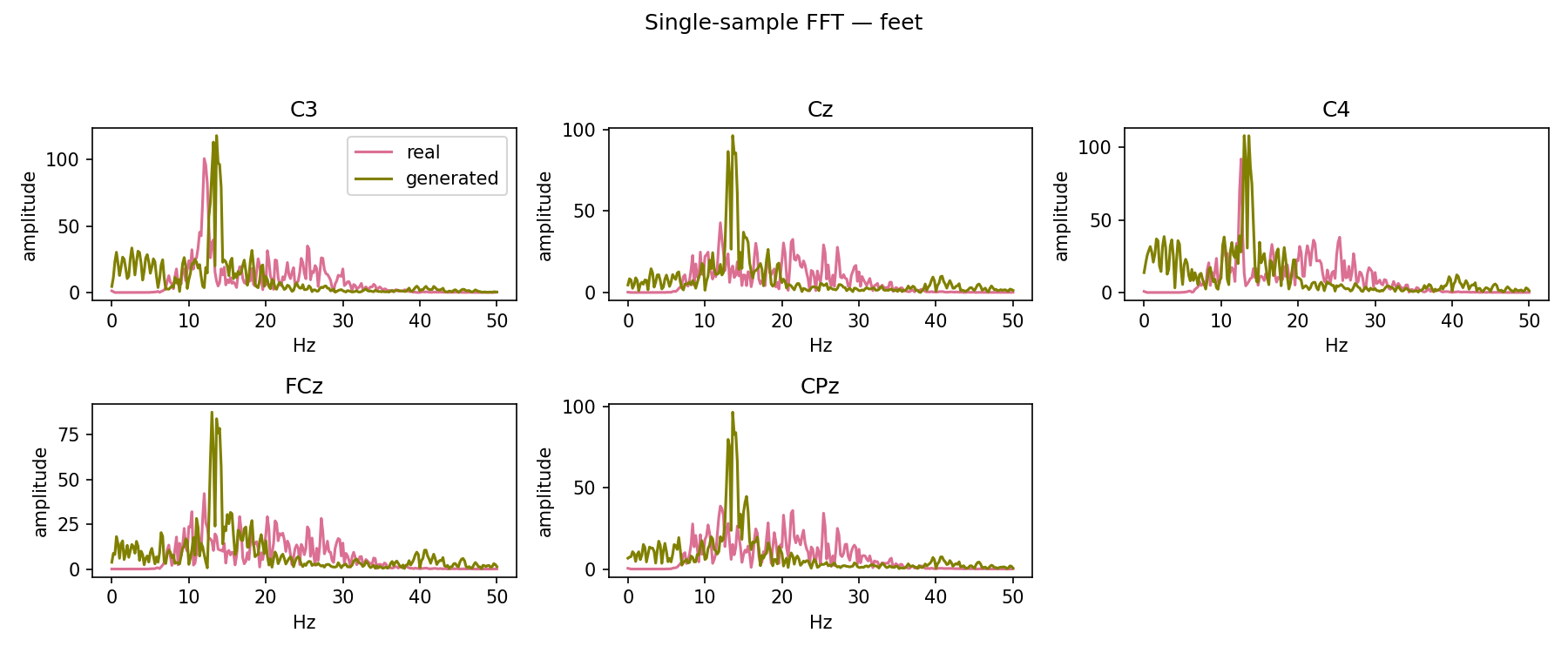}
\caption{Single-sample FFT (one real feet trial vs.\ one generated feet sample; per channel; not averaged).}
\label{fig:fft-single}
\end{figure}

\begin{figure}[htbp]
\centering
\includegraphics[width=0.85\textwidth]{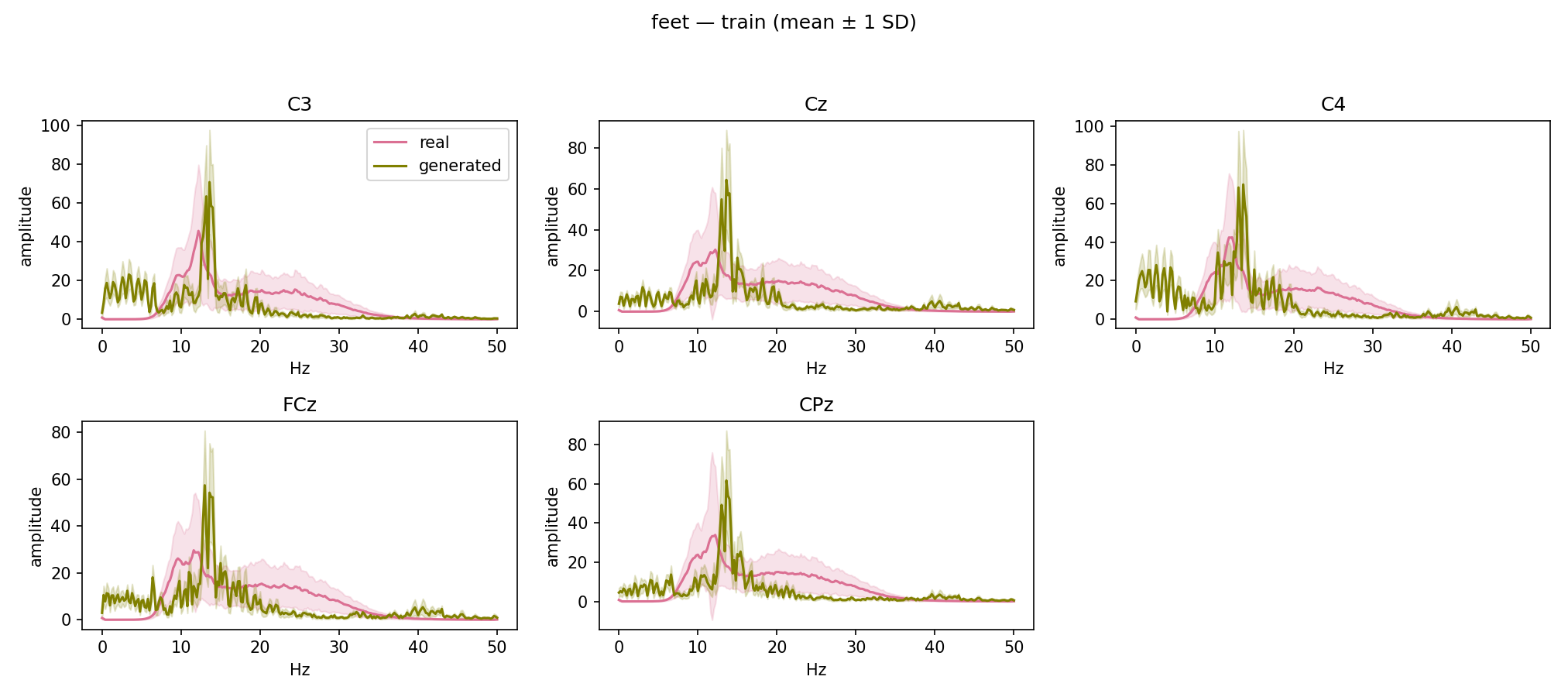}
\caption{Trial-averaged spectra with variability shading, real vs.\ generated --- feet (one panel per channel; solid line = mean across trials; shaded band = $\pm$1 SD; real in pink, generated in olive).}
\label{fig:fft-trial-avg}
\end{figure}

The frequency-domain analysis backs up the conclusions drawn from the signal time representation. Motor imagery is characterised by power in the 8--15\,Hz sensorimotor (mu) band. Looking at the single-trial FFT of the real vs.\ reconstructed signal (Figure~\ref{fig:fft-single}), we see similar behaviour, with power in the mu band, though the reconstructed signal carries more artefacts in the lower-frequency range. However, once the signals are averaged per class, the mu-band power is reconstructed, but the spectral shape and variability differ substantially from the real data (Figure~\ref{fig:fft-trial-avg}).

\subsubsection{Topographic Maps}

Topographic maps of per-channel variance, evaluating the spatial covariance structure of each class, are also shown. With only five channels, the maps are heavily interpolated, so they are read for gross layout (overall scale, and left--right / front-central gradient orientation) rather than fine spatial detail.

\begin{figure}[htbp]
\centering
\includegraphics[width=0.85\textwidth]{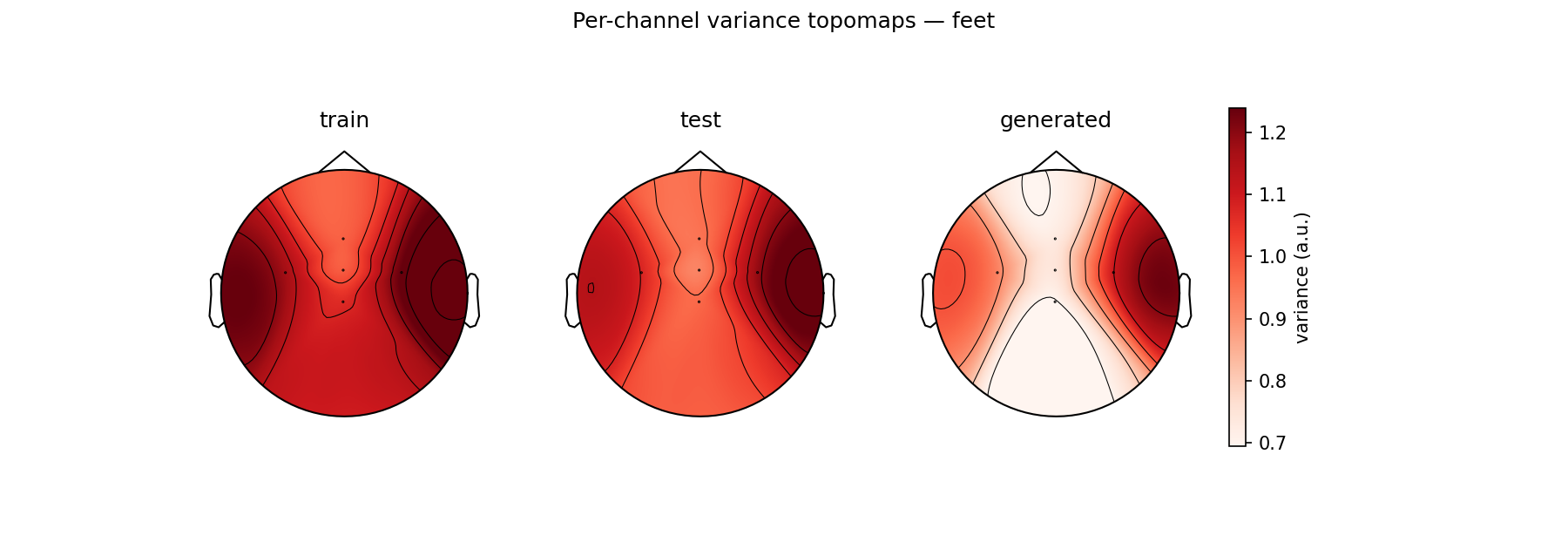}
\caption{Per-channel variance topomaps, feet (columns: train/test/generated). Train-vs-test correlation 0.971 (real data's own split-to-split consistency), train-vs-generated correlation 0.909, generated total variance = 77\% of train.}
\label{fig:topomap-feet}
\end{figure}

\begin{figure}[htbp]
\centering
\includegraphics[width=0.85\textwidth]{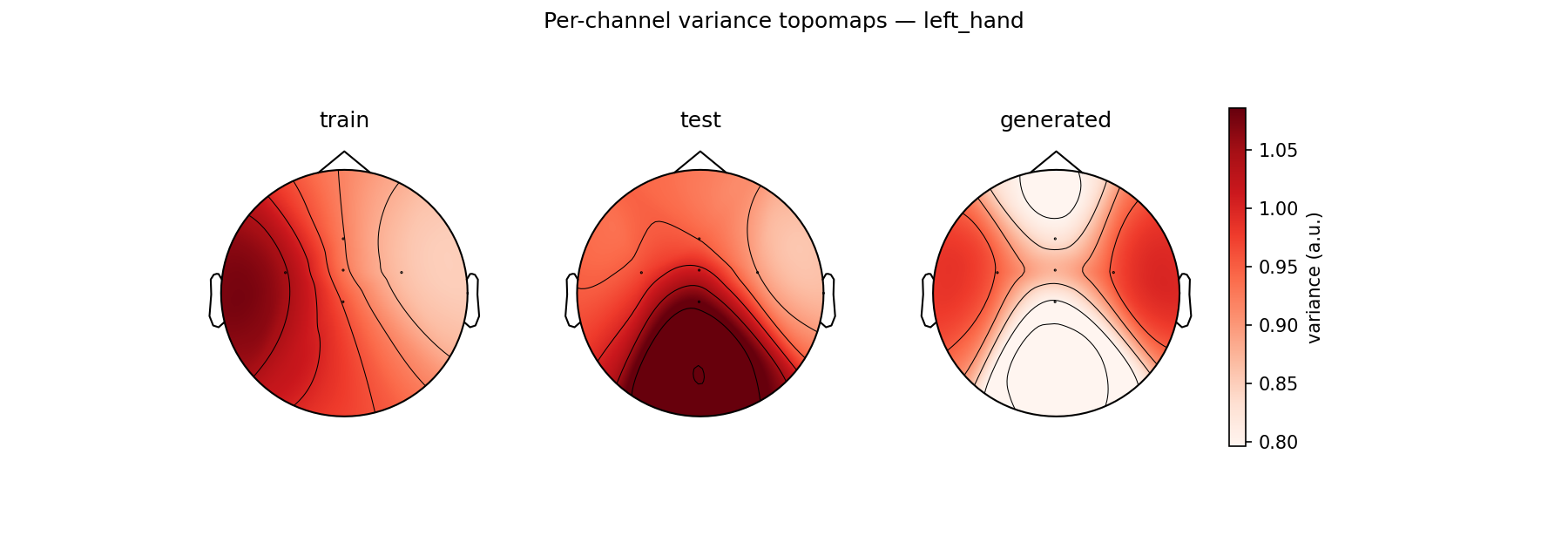}
\caption{Per-channel variance topomaps, left\_hand (columns: train/test/generated). Train-vs-test correlation is only 0.379 here --- well below feet's 0.971 --- meaning even the real data's own train/test split is spatially inconsistent for this class, before the generator is involved at all. Against that noisy target, train-vs-generated correlation is effectively zero (0.032): the generated left\_hand topography does not track the real spatial pattern, though generated total variance (93\% of train) is close to the real scale.}
\label{fig:topomap-lefthand}
\end{figure}

\begin{figure}[htbp]
\centering
\includegraphics[width=0.85\textwidth]{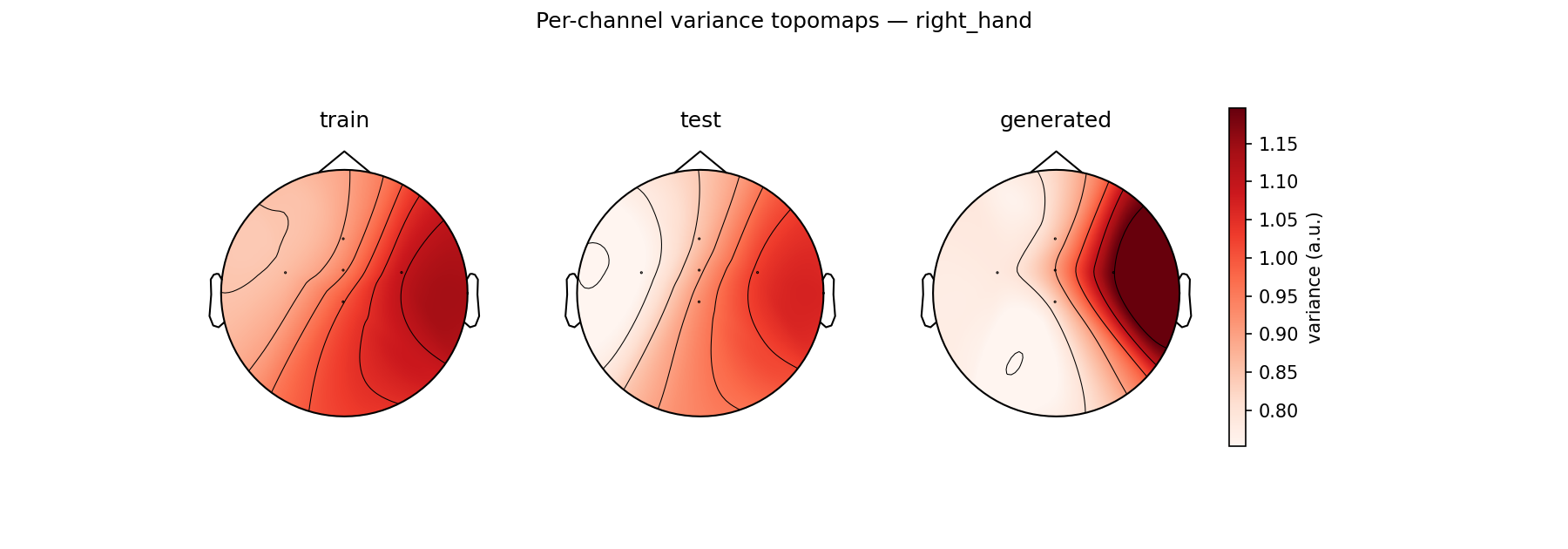}
\caption{Per-channel variance topomaps, right\_hand (columns: train/test/generated). Train-vs-test correlation 0.976 and train-vs-generated correlation 0.865 --- both close to feet's numbers, and generated total variance (95\% of train) is the closest to real of the three classes.}
\label{fig:topomap-righthand}
\end{figure}

Spatial-covariance fidelity is therefore not uniform across classes: feet and right\_hand both show a real, well-preserved train/test pattern that the generator tracks reasonably well, but left\_hand's own train/test split is barely self-consistent and the generator's left\_hand topography does not track it at all. The feet-only figures shown earlier in this section should not be read as representative of every class; left\_hand in particular is a materially weaker case for this model. Generated maps otherwise show broadly similar behaviour to the train data on the two better-behaved classes, consistent with the covariance constraint improving part of the spatial fidelity of the generated data.

\section{Discussion}
\label{sec:discussion}

The \textbf{Aug-only} results show that the generator has learned meaningful class structure: synthetic data alone is sufficient to train covariance-based and latent-space classifiers well above chance, in some cases approaching the accuracy achieved using real data. This indicates that the class-conditional prior captures genuine class-discriminative structure rather than producing generic or unstructured noise.

However, the same synthetic data provides little benefit when added to real training data. If the generated trials captured variability absent from the original dataset, training on both real and synthetic data would be expected to outperform training on real data alone. Instead, the with-Aug results remain close to the \textbf{Train} baseline across almost all classifiers and under both evaluation protocols, suggesting that the generated samples contribute little additional information.

The reconstruction analysis points to one possible explanation. Individual generated trials do not resemble their corresponding real waveforms in a sample-by-sample sense, instead reproducing only the broader spectral and covariance characteristics of each class. Consequently, classifiers that rely on the temporal structure of the EEG signal, such as EEGNet, have little to gain from these synthetic trials and may even be degraded by them, since the generated signals resemble class prototypes more than realistic individual observations.

However, the latent-space results suggest that decoder fidelity is not the only limiting factor. Even when classification is performed directly on the encoder's latent representation --- bypassing the decoder entirely --- augmentation produces little improvement over training on real data alone. This indicates that, although the latent representation is discriminative, it contributes little information beyond what is already captured by the real training set. Consequently, even a substantially better decoder would not necessarily be expected to produce large gains in downstream classification accuracy.

\section{Limitations}
\label{sec:limitations}

This study has several limitations that bound how broadly the results should be read. First, all results are drawn from a single dataset, Zhou2016, a single three-class motor-imagery dataset; none of the findings here have been tested on a second dataset, a different number of classes, or a different recording setup. Second, sample sizes are small: the cross-seed study uses three seeds, the cross-user (LOSO) study uses four subjects/folds, and the data-quality section uses a single within-user model for visual and fidelity inspection. Third, the study is likely underpowered for some comparisons; readers should not treat a confidence interval's failure to exclude zero as evidence that an effect is absent, since a larger study, with more seeds and more subjects, would be needed to resolve this either way. Finally, the fidelity analyses are partly qualitative: the reconstruction, spectral, and topographic panels each report quantitative numbers computed from the saved trial data, but none constitutes a full diversity metric capturing how much novel variability the generated trials add relative to the training distribution, an assessment made difficult by the fact that the definition of what good EEG data looks like is itself hard to pin down.

\section{Conclusion}
\label{sec:conclusion}

We trained a class-conditional VAE with an integrated latent classifier and a cycle-consistency-refined decoder on motor-imagery EEG, and evaluated its usefulness as a data-augmentation source for four independent downstream classifiers under both within-user and cross-user (LOSO) protocols. The generator learns a latent space with genuine class structure --- nearest-prior-mean and latent-classifier accuracy both approach signal-space classifier accuracy, and Aug-only training reaches well above chance for covariance-based pipelines. This structure, however, does not translate into a consistent improvement when synthetic trials are added to real training data: with-Aug deltas are small and not statistically distinguishable from zero for CSP+LDA, TGSP+SVM, and EEGNet in both protocols, with MDM the sole classifier showing a consistent (if modest, and not cross-user-transferable) positive point estimate. The ablation studies (Appendix~\ref{app:ablation}) indicate that the Log-Euclidean covariance constraint is essential for Aug-only usability (its removal collapses synthetic-only accuracy to near chance), that Soft-DTW recovers physiologically plausible mu-band content that MSE reconstruction does not, and that the alternating decoder-focused training schedule improves MDM's usable signal without harming other classifiers. Taken together, these results support treating this class of CVAE as a tool for producing class-structured, covariance-consistent synthetic EEG --- useful for probing what a downstream pipeline can extract from class-level structure alone --- rather than as a general-purpose substitute or supplement for real recordings. Future work should test these conclusions on additional datasets and class counts, and should pursue diversity metrics that can distinguish ``faithful class prototype'' generation from genuinely novel, informative synthetic variability.

\section*{Acknowledgments}

This work was supported by the French National Research Agency (ANR), with projects PROTEUS (grant ANR-22-CE33-0015-01).

\section*{AI/LLM Usage Disclosure}

Large language model (AI) tools were used to assist with drafting and editing portions of this manuscript's text, and with development, debugging, and refactoring of portions of the accompanying analysis and training code. All AI-assisted text and code were reviewed, verified, and edited by the authors, who take full responsibility for the accuracy, integrity, and conclusions of the work.

\bibliographystyle{unsrtnat}
\bibliography{references}

@article{zhou2016fully,
  author  = {Zhou, Bin and Wu, Xiaopei and Lv, Zhen and Zhang, Lai and Guo, Xin},
  title   = {A fully automated trial selection method for optimization of motor imagery based brain-computer interface},
  journal = {PLOS ONE},
  volume  = {11},
  number  = {9},
  pages   = {e0162657},
  year    = {2016},
  doi     = {10.1371/journal.pone.0162657}
}

@inproceedings{sohn2015learning,
  author    = {Sohn, Kihyuk and Yan, Xinchen and Lee, Honglak},
  title     = {Learning structured output representation using deep conditional generative models},
  booktitle = {Advances in Neural Information Processing Systems (NeurIPS)},
  volume    = {28},
  pages     = {3483--3491},
  year      = {2015}
}

@misc{polaka2026riemannian,
  author       = {Po\c{l}aka, Valters and de Jong, Ivo P. and Sburlea, Andreea I.},
  title        = {Riemannian geometry-preserving variational autoencoder for {MI-BCI} data augmentation},
  year         = {2026},
  eprint       = {2603.10563},
  archivePrefix = {arXiv}
}

@inproceedings{bethge2022eeg2vec,
  author    = {Bethge, David and Hallgarten, Philipp and Grosse-Puppendahl, Tobias and Kari, Mohamed and Chuang, Lewis L. and \"{O}zdenizci, Ozan and Schmidt, Albrecht},
  title     = {{EEG2Vec}: Learning affective {EEG} representations via variational autoencoders},
  booktitle = {Proc. 2022 IEEE Int. Conf. Systems, Man, and Cybernetics (SMC)},
  pages     = {3150--3157},
  year      = {2022},
  doi       = {10.1109/SMC53654.2022.9945517}
}

@incollection{zancanaro2024veegnet,
  author    = {Zancanaro, Alberto and Cisotto, Giulia and Zoppis, Italo and Manzoni, Sara L.},
  title     = {{vEEGNet}: Learning latent representations to reconstruct {EEG} raw data via variational autoencoders},
  booktitle = {Information and Communication Technologies for Ageing Well and e-Health},
  series    = {Communications in Computer and Information Science},
  publisher = {Springer},
  pages     = {114--129},
  year      = {2024},
  doi       = {10.1007/978-3-031-62753-8_7}
}

@article{cisotto2024hveegnet,
  author  = {Cisotto, Giulia and Zancanaro, Alberto and Zoppis, Italo F. and Manzoni, Sara L.},
  title   = {{hvEEGNet}: A novel deep learning model for high-fidelity {EEG} reconstruction},
  journal = {Frontiers in Neuroinformatics},
  volume  = {18},
  pages   = {1459970},
  year    = {2024},
  doi     = {10.3389/fninf.2024.1459970}
}

@article{zhao2024vaeeg,
  author  = {Zhao, Taorong and Cui, Yuxin and Ji, Ting and Luo, Jinyuan and Li, Wenlong and Jiang, Jian and Gao, Zheng and Hu, Wei and Yan, Yueming and Jiang, Yuwu and Hong, Bo},
  title   = {{VAEEG}: Variational auto-encoder for extracting {EEG} representation},
  journal = {NeuroImage},
  volume  = {304},
  pages   = {120946},
  year    = {2024},
  doi     = {10.1016/j.neuroimage.2024.120946}
}

@article{doku2026structure,
  author  = {Doku, Michael and Tibermacine, Imad Eddine and Russo, Stefania and Rabehi, Abdelkader and Habib, Maad and Napoli, Christian},
  title   = {Structure-preserving {EEG} augmentation via {R}iemannian conditional generative adversarial networks},
  journal = {IEEE Access},
  year    = {2026},
  doi     = {10.1109/ACCESS.2026.3663245}
}

@article{williams2025eeggan,
  author  = {Williams, Christoph C. and Weinhardt, Damian and Hewson, Jasper and P{\l}omecka, Martyna B. and Langer, Nicolas and Musslick, Sebastian},
  title   = {{EEG-GAN}: A generative {EEG} augmentation toolkit for enhancing neural classification},
  journal = {bioRxiv},
  year    = {2025},
  doi     = {10.1101/2025.06.23.661164}
}

@article{yao2025transcvaegan,
  author  = {Yao, Yuang and Wang, Xin and Hao, Xiaoyu and Sun, Han and Dong, Rui and Li, Yang},
  title   = {{Trans-cVAE-GAN}: Transformer-based {cVAE-GAN} for high-fidelity {EEG} signal generation},
  journal = {Bioengineering},
  volume  = {12},
  number  = {10},
  pages   = {1028},
  year    = {2025},
  doi     = {10.3390/bioengineering12101028}
}

@inproceedings{cuturi2017softdtw,
  author    = {Cuturi, Marco and Blondel, Mathieu},
  title     = {Soft-{DTW}: a differentiable loss function for time-series},
  booktitle = {Proc. 34th Int. Conf. Machine Learning (ICML)},
  series    = {PMLR},
  volume    = {70},
  pages     = {894--903},
  year      = {2017},
  note      = {arXiv:1703.01541}
}

@inproceedings{jha2018disentangling,
  author    = {Jha, Ananya Harsh and Anand, Saket and Singh, Maneesh and Veeravasarapu, V. S. R.},
  title     = {Disentangling factors of variation with cycle-consistent variational auto-encoders},
  booktitle = {Computer Vision -- ECCV 2018},
  series    = {Lecture Notes in Computer Science},
  volume    = {11207},
  publisher = {Springer},
  pages     = {829--845},
  year      = {2018},
  doi       = {10.1007/978-3-030-01219-9_49}
}

@misc{silvestri2025covae,
  author       = {Silvestri, Gabriele and Ambrogioni, Luca},
  title        = {{CoVAE}: Consistency training of variational autoencoders},
  year         = {2025},
  eprint       = {2507.09103},
  archivePrefix = {arXiv}
}

@misc{caillon2021rave,
  author       = {Caillon, Antoine and Esling, Philippe},
  title        = {{RAVE}: A variational autoencoder for fast and high-quality neural audio synthesis},
  year         = {2021},
  eprint       = {2111.05011},
  archivePrefix = {arXiv}
}

@article{vandermaaten2008visualizing,
  author  = {van der Maaten, Laurens and Hinton, Geoffrey},
  title   = {Visualizing data using t-{SNE}},
  journal = {Journal of Machine Learning Research},
  volume  = {9},
  number  = {86},
  pages   = {2579--2605},
  year    = {2008}
}
\newpage
\appendix

\section{Appendix A: Ablation Studies}
\label{app:ablation}

These single-factor studies isolate the contribution of individual design choices; each varies only the factor under study with the reported configuration otherwise fixed, and downstream augmentation is scored on the held-out test split. Section~\ref{app:schedule} compares the proposed alternating training schedule against standard end-to-end VAE training to evaluate whether the decoder-focused phase improves the usefulness of generated samples. Section~\ref{app:reconloss} compares Soft-DTW and mean squared error (MSE) as the reconstruction objective to assess their effect on reconstruction fidelity and downstream augmentation performance. Section~\ref{app:classsep} investigates the relative contribution of the conditional KL divergence and the latent classifier to learning class-discriminative latent representations. Section~\ref{app:covariance} evaluates the impact of removing the Log-Euclidean covariance loss to determine its importance for preserving class-discriminative signal structure. Section~\ref{app:lossbalance} scales the reconstruction and covariance objectives relative to the latent regularisation and classification objectives to study the trade-off between signal fidelity and latent discriminability. Section~\ref{app:klratio} varies the relative weighting between the KL divergence and latent classification losses to examine how strongly latent regularisation should influence class separation. Section~\ref{app:covratio} varies the covariance-loss weight while keeping the reconstruction loss fixed to study the balance between temporal reconstruction fidelity and preservation of covariance structure.

The ablation studies share the following base configuration:

\begin{table}[H]
\centering
\caption{Shared ablation-study base configuration.}
\label{tab:t13}
\small
\begin{tabular}{@{}l p{11cm}@{}}
\toprule
Group & Setting \\
\midrule
Data & Zhou 2016~\citep{zhou2016fully}, 3 classes, 5 channels, 8--32\,Hz, 1--6\,s, split 60/20/20 (grouped within subject/session), z-scored \\
Architecture & latent dim 100, 1 encoder + 1 convolutional decoder, trainable class prior, classifier hidden width 64 \\
Objective (base) & Soft-DTW recon (0.01), conditional KL (5.0), Log-Euclidean covariance (1.5), cross-entropy (1.0) \\
Optimisation & Adam, lr 1e-3, batch size 30 \\
Evaluation & CSP+LDA, TGSP+SVM, MDM, EEGNet; 95\% t-CIs; 3 seeds for multi-seed studies \\
\bottomrule
\end{tabular}
\end{table}

\subsection{Training Schedule: Alternating vs.\ Normal}
\label{app:schedule}

Comparing an alternating schedule (180 normal + 180 decoder-focused epochs) vs.\ a normal-only schedule (360 epochs). This tests whether the decoder-focused phase is worth its cost.

\begin{table}[H]
\centering
\caption{Alternating vs.\ normal training.}
\label{tab:t14}
\small
\begin{tabular}{@{}lccc@{}}
\toprule
Classifier / condition & Normal & Alternating & $\Delta$ (alt $-$ normal) \\
\midrule
CSP+LDA --- with Aug & 0.761 & 0.786 & +0.025 \\
TGSP+SVM --- with Aug & 0.786 & 0.789 & +0.003 \\
MDM --- with Aug & 0.619 & 0.697 & +0.078 \\
\bottomrule
\end{tabular}
\end{table}

The decoder-focused phase brings a large improvement to the MDM classification performance, even if this effect is not shared by the other 2 classifiers.

\subsection{Reconstruction Loss --- Soft-DTW vs.\ MSE}
\label{app:reconloss}

Comparing MSE reconstruction (weight 1.0) vs.\ Soft-DTW (weight 0.01), all else matched (conditional KL 5.0, Log-Euclidean covariance 1.5), normal schedule, 200 epochs. The two weights differ by 100$\times$ because the two losses sit on very different natural scales.

\begin{table}[H]
\centering
\caption{Reconstruction loss impact on test data accuracy.}
\label{tab:t15}
\footnotesize
\begin{tabular}{@{}p{2.1cm}*{2}{>{\centering\arraybackslash}p{1.4cm}}*{3}{>{\centering\arraybackslash}p{2.5cm}}@{}}
\toprule
Reconstruction loss & Latent acc & Silhouette & CSP+LDA Train / with-Aug & TGSP+SVM Train / with-Aug & MDM Train / with-Aug \\
\midrule
MSE & 0.803 & 0.171 & 0.767 / 0.772 & 0.794 / 0.786 & 0.633 / 0.619 \\
Soft-DTW & 0.803 & 0.138 & 0.767 / 0.763 & 0.794 / 0.777 & 0.633 / 0.600 \\
\bottomrule
\end{tabular}
\end{table}

The table alone doesn't show \emph{why} the reconstruction loss choice matters --- both losses give the VAE's encoder-side latent structure essentially the same downstream accuracy, but they produce visibly different decoder outputs.

\begin{figure}[htbp]
\centering
\includegraphics[width=0.85\textwidth]{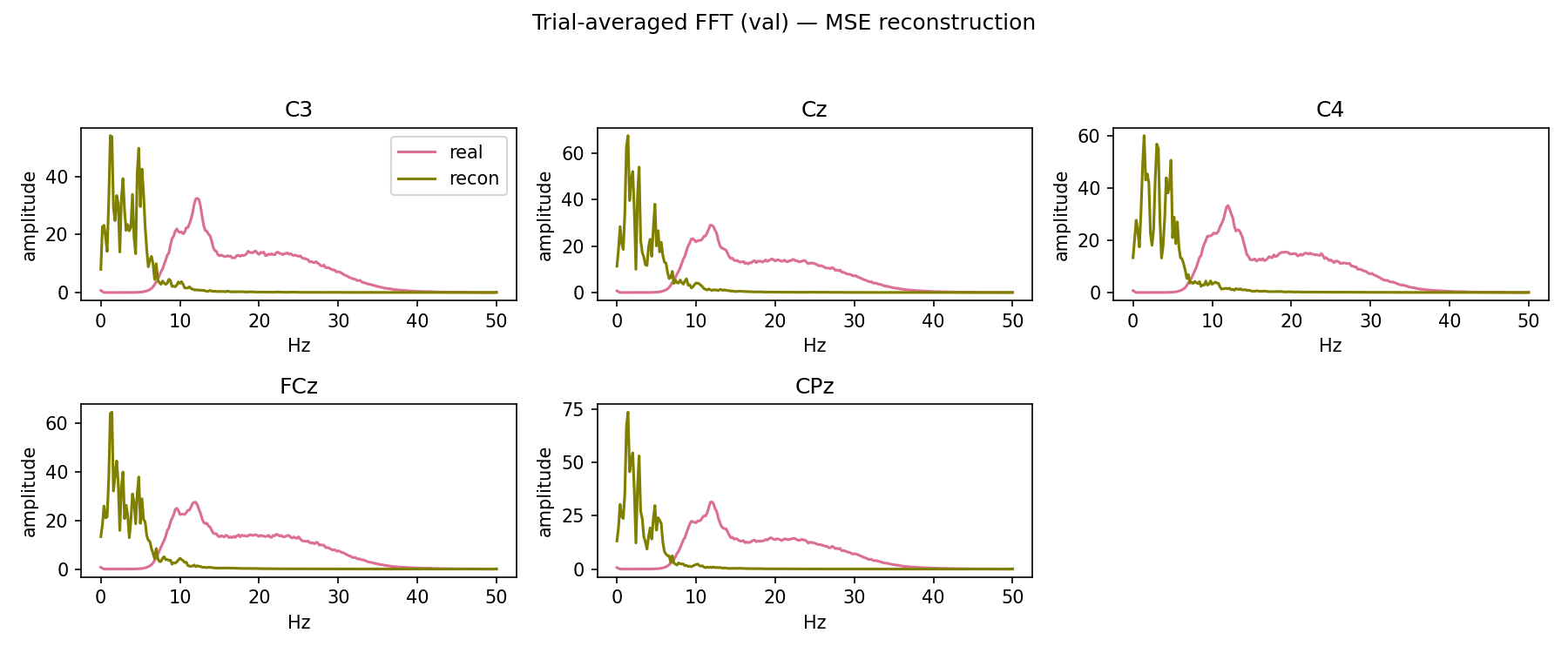}\\[0.5em]
\includegraphics[width=0.85\textwidth]{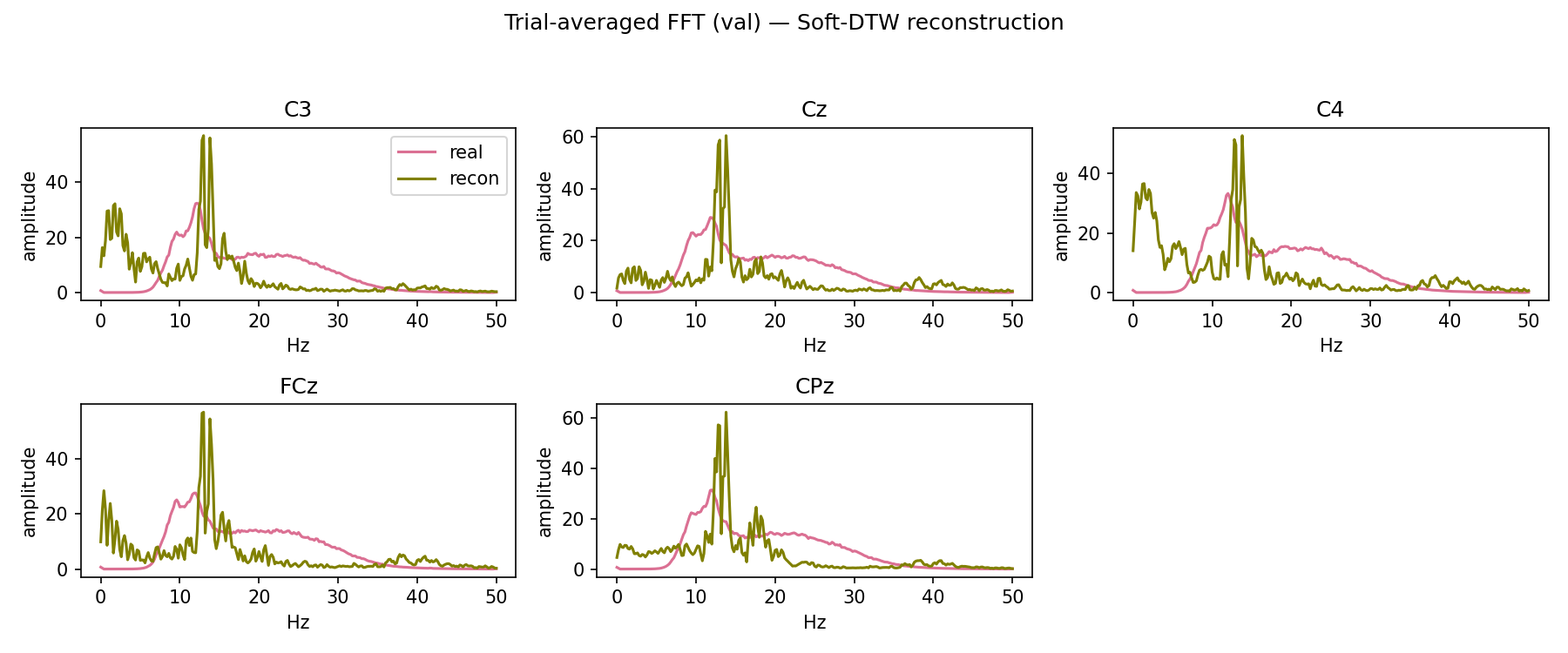}
\caption{Trial-averaged FFT amplitude, real (pink) vs.\ reconstruction (olive), MSE (top) vs.\ Soft-DTW (bottom), all 5 channels.}
\label{fig:fft-mse-dtw}
\end{figure}

This makes the difference precise: MSE's reconstruction spectrum peaks near 0--5\,Hz, with little to no power in the real mu-band peak ($\approx$10--15\,Hz). Soft-DTW's reconstruction spectrum peaks in the same $\approx$10--15\,Hz band the real data does (sharper and taller than the real peak, plus a little spurious energy near 35--40\,Hz, but in the right place), correctly identifying where the physiologically relevant content actually is.

The interpretation is that Soft-DTW's phase-tolerance lets the decoder match the mu-rhythm's frequency content and rough temporal structure without being forced to align every sample exactly; MSE, forced to minimise a sample-wise squared error against a partially-misaligned target, ``hedges'' toward a smoothed near-DC signal that minimises average error at the cost of reproducing the signal's actual rhythm. So although MSE and Soft-DTW are statistically indistinguishable on this ablation's downstream classifiers, Soft-DTW is the only one of the two that produces a reconstruction anyone would recognise as motor-imagery EEG. Soft-DTW is retained for generative fidelity on that basis, not because it wins on this table.

\subsection{Source of Class Separation --- KL vs.\ KL and Classifier (Both)}
\label{app:classsep}

Comparing two variants: KL-only (classifier off), both (reported) using normal schedule, 1 seed, 200 epochs.

\begin{table}[H]
\centering
\caption{Class-separation variants test accuracy on different classification algorithms.}
\label{tab:t16}
\footnotesize
\begin{tabular}{@{}p{2.6cm}>{\centering\arraybackslash}p{1.8cm}*{3}{>{\centering\arraybackslash}p{2.8cm}}@{}}
\toprule
Variant & Nearest prior mean & CSP+LDA Baseline / with-Aug & TGSP+SVM Baseline / with-Aug & MDM Baseline / with-Aug \\
\midrule
KL-only (no classifier) & 0.719 & 0.767 / 0.769 & 0.794 / 0.756 & 0.633 / 0.586 \\
Both (reported) & 0.839 & 0.767 / 0.775 & 0.794 / 0.767 & 0.633 / 0.603 \\
\bottomrule
\end{tabular}
\end{table}

The nearest-prior-mean rule shows that using both the KL term and the classifier leads to better latent-space separation, even though this does not translate into better downstream classifier performance. The large variation seen in classification accuracy makes it hard to say how impactful the classifier term really is, but since it does not degrade performance, it is kept.

\begin{figure}[htbp]
\centering
\includegraphics[width=0.7\textwidth]{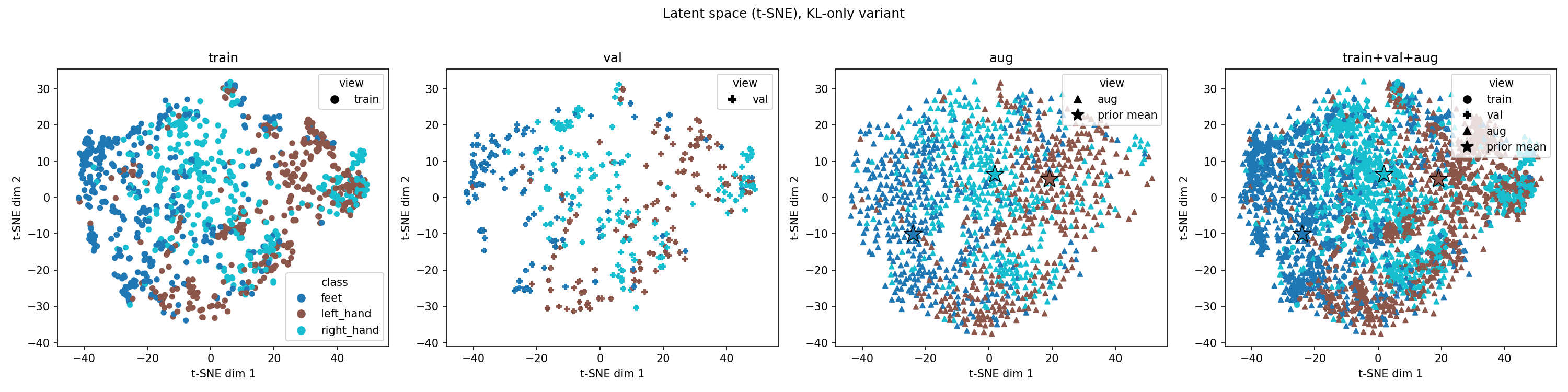}
\caption{Latent space (t-SNE), KL-only variant. Same construction as Figure~\ref{fig:tsne} (train / val+test / aug / all, class-conditional prior means as stars); classifier head disabled, so class structure here comes entirely from the KL term.}
\label{fig:tsne-kl-only}
\end{figure}

\begin{figure}[htbp]
\centering
\includegraphics[width=0.7\textwidth]{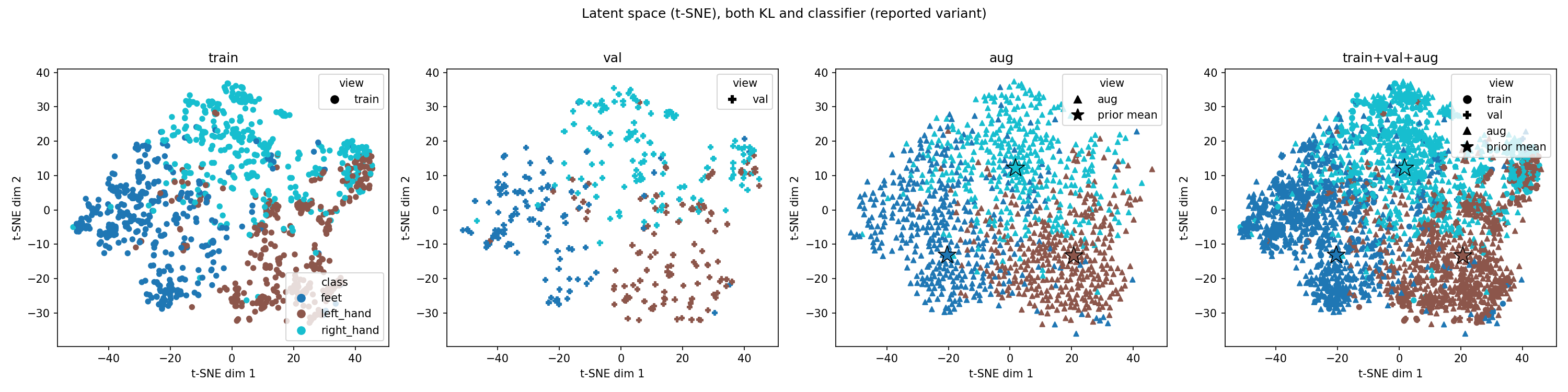}
\caption{Latent space (t-SNE), both KL and classifier (reported variant).}
\label{fig:tsne-both}
\end{figure}

Visually, both variants show three separable class clusters with the prior means (stars) landing inside their respective clusters --- consistent with the KL-only nearest-prior-mean accuracy (0.719) already being well above chance without any classifier loss at all. The ``both'' variant's clusters are not obviously tighter or better-separated by eye than KL-only's, matching Table~\ref{tab:t16}'s numeric finding that adding the classifier loss on top of the KL term does not visibly change the latent geometry, even though it does raise nearest-prior-mean accuracy (0.719 $\to$ 0.839).

\subsection{Covariance Constraint --- On vs.\ Off}
\label{app:covariance}

Looking at the impact for Log-Euclidean covariance, everything else matched. Normal schedule, 1 seed, 200 epochs.

\begin{table}[H]
\centering
\caption{Covariance constraint on/off results on test accuracy.}
\label{tab:t17}
\footnotesize
\begin{tabular}{@{}p{2.9cm}>{\centering\arraybackslash}p{1.8cm}*{3}{>{\centering\arraybackslash}p{2.5cm}}@{}}
\toprule
Variant & Nearest prior mean & CSP+LDA Train / Aug & TGSP+SVM Train / Aug & MDM Train / Aug \\
\midrule
No covariance & 0.347 & 0.767 / 0.297 & 0.794 / 0.297 & 0.633 / 0.297 \\
Log-Euclidean (reported) & 0.800 & 0.767 / 0.675 & 0.794 / 0.706 & 0.633 / 0.553 \\
\bottomrule
\end{tabular}
\end{table}

The effect of the covariance constraint becomes most apparent when classifiers are trained exclusively on synthetic data. Without the Log-Euclidean covariance constraint, the model fails to learn meaningful covariance relationships between channels, causing synthetic-only classification performance to collapse. While this is not entirely surprising given the difficulty of reconstructing realistic EEG signals, it is notable that the collapse also extends to the latent-space classifier. This suggests that the covariance constraint influences not only the generated signals but also the representations learned by the encoder. One possible interpretation is that the encoder's latent representation derives much of its discriminative power from covariance information encouraged by the covariance loss. Although this is desirable for covariance-based downstream classifiers, the collapse of latent-space performance when the covariance constraint is removed suggests that the encoder is learning little class-discriminative information beyond covariance.

\subsection{Loss-Weight Balance --- Data-Quality vs.\ Class-Separation Terms}
\label{app:lossbalance}

Comparing different trade-offs between data quality and class separation. The data-quality group (reconstruction + covariance) is scaled by a multiplier $\lambda$ while the class-separation group (KL 5.0 + classification 1.0) is fixed. $\lambda \in \{0.25, 0.5, 1, 2, 4\}$; $\lambda=1$ is the reported model.

\begin{table}[H]
\centering
\caption{Data-quality weight $\lambda$ impact.}
\label{tab:t18}
\small
\begin{tabular}{@{}>{\centering\arraybackslash}p{1.2cm}>{\centering\arraybackslash}p{1.8cm}>{\centering\arraybackslash}p{2.8cm}>{\centering\arraybackslash}p{2.8cm}>{\centering\arraybackslash}p{2.2cm}@{}}
\toprule
$\lambda$ & Nearest prior mean & Log-Euclidean distance on test data & Avg Soft-DTW (test, paired) & MDM +aug / aug \\
\midrule
0.25 & 0.828 & 1.024 & 891.5 & 0.625 / 0.561 \\
0.5 & 0.839 & 0.880 & 866.0 & 0.619 / 0.559 \\
1.0 (reported) & 0.792 & 0.726 & 839.5 & 0.623 / 0.564 \\
2.0 & 0.783 & 0.635 & 849.2 & 0.627 / 0.575 \\
4.0 & 0.769 & 0.669 & 827.7 & 0.626 / 0.567 \\
\bottomrule
\end{tabular}
\end{table}

As $\lambda$ rises, test-set reconstruction fidelity improves on both metrics overall: covariance distance and average Soft-DTW distance fall --- though neither is perfectly monotonic. Class separation moves in the opposite direction: the unfitted nearest-prior-mean rule peaks at $\lambda=0.5$ (\textbf{0.839}), then declines steadily as $\lambda$ keeps rising (0.792 $\to$ 0.783 $\to$ 0.769). This is the same data-fidelity-vs-class-separation trade-off that was this ablation study's central hypothesis. Downstream augmentation accuracy is insensitive to $\lambda$ across the whole range. The choice of $\lambda=1$ remains a reasonable middle ground.

\subsection{KL/Classification Weight Ratio}
\label{app:klratio}

The study extends Section~\ref{app:classsep}'s toggle (class+KL / KL-only) into a 6-point curve: class\_weight fixed at 1.0, cov\_weight fixed at 1.5, recon\_mode="soft-dtw" fixed at 0.01, and kl\_weight swept over $\{0, 0.5, 1, 2.5, 5, 10\}$ (kl\_weight=5.0 is the reported model). Normal schedule, 1 seed, 200 epochs.

\begin{table}[H]
\centering
\caption{KL/classification ratio impact on test accuracy.}
\label{tab:t19}
\small
\begin{tabular}{@{}>{\centering\arraybackslash}p{2cm}>{\centering\arraybackslash}p{2.8cm}>{\centering\arraybackslash}p{3.5cm}@{}}
\toprule
KL weight & Nearest prior mean & TGSP+SVM Baseline / with-Aug \\
\midrule
0 (no KL) & 0.297 & 0.794 / 0.767 \\
0.5 & 0.797 & 0.794 / 0.767 \\
1 & 0.808 & 0.794 / 0.767 \\
2.5 & 0.817 & 0.794 / 0.781 \\
5.0 (reported) & 0.831 & 0.794 / 0.764 \\
10 & 0.781 & 0.794 / 0.769 \\
\bottomrule
\end{tabular}
\end{table}

The impact of the KL weighting on the downstream classifiers is negligible. The weighting with the best nearest-prior-mean performance is KL=5, and is therefore chosen as the reported configuration; however, the difference from neighbouring KL weights is not significant given the variation seen from data partitioning alone.

\subsection{Reconstruction/Covariance Weight Ratio}
\label{app:covratio}

Extends Section~\ref{app:covariance}'s on/off toggle (no-covariance / Log-Euclidean) into a 6-point curve: recon\_mode="soft-dtw" at 0.01 and kl\_weight fixed at 5.0 throughout, class\_weight fixed at 1.0, and cov\_weight swept over $\{0, 0.5, 1, 1.5, 3, 5\}$ (cov\_weight=1.5 is the reported model).

\begin{table}[H]
\centering
\caption{Reconstruction/covariance ratio's impact on test accuracy.}
\label{tab:t20}
\footnotesize
\begin{tabular}{@{}*{2}{>{\centering\arraybackslash}p{1.7cm}}*{2}{>{\centering\arraybackslash}p{2.2cm}}*{2}{>{\centering\arraybackslash}p{2.4cm}}@{}}
\toprule
Covariance weight & Nearest prior mean & Covariance distance (log-Euclidean, test) & Avg Soft-DTW (test, paired) & TGSP+SVM Baseline / with-Aug & MDM Baseline / with-Aug \\
\midrule
0 (no covariance) & 0.364 & 84.383 & 1026.5 & 0.794 / 0.792 & 0.633 / 0.297 \\
0.5 & 0.836 & 1.757 & 1141.3 & 0.794 / 0.783 & 0.633 / 0.536 \\
1.0 & 0.803 & 1.104 & 1179.3 & 0.794 / 0.761 & 0.633 / 0.564 \\
1.5 (reported) & 0.806 & 0.982 & 1185.9 & 0.794 / 0.764 & 0.633 / 0.589 \\
3.0 & 0.808 & 0.803 & 1238.1 & 0.794 / 0.767 & 0.633 / 0.617 \\
5.0 & 0.775 & 0.858 & 1249.2 & 0.794 / 0.772 & 0.633 / 0.625 \\
\bottomrule
\end{tabular}
\end{table}

As expected, the weighting acts as a trade-off between reconstruction fidelity and covariance fidelity. Despite this, the trade-off is not visible in the downstream classifiers outside of MDM, and even for MDM the improvement is small compared to the variation observed in classification results generally. We pick cov\_weight=1.5 as a middle ground between reconstruction and covariance fidelity.

\section{Appendix B: Per-Subject/Seed Augmentation Results}
\label{app:persubject}

Section~\ref{sec:results} reports augmentation impact aggregated across seeds (within-user) or subjects (cross-user), which is the right view for the paper's headline claims but hides how much any one seed or subject varies on its own. This appendix breaks the same results down \textbf{per seed} (Section~\ref{app:withinuser-perseed}) and \textbf{per held-out user} (Section~\ref{app:crossuser-persubject}), with the draw-to-draw variation shown alongside each point estimate, so a reader can see directly whether a given seed's or subject's own result is a stable estimate or a noisy one before it gets folded into the cross-seed/cross-subject aggregate. This is the more in-depth, single-partition view; the aggregated view in Section~\ref{sec:results} remains the primary evidence for the paper's claims.

\subsection{Within-User (Per-Seed)}
\label{app:withinuser-perseed}

\begin{table}[H]
\centering
\caption{Within-user augmentation, per seed --- Seed 0.}
\label{tab:t21-s0}
\small
\begin{tabular}{@{}lccc@{}}
\toprule
Classifier & Baseline & with Aug & Aug only \\
\midrule
CSP+LDA & 0.769 & 0.785 $\pm$ 0.016 & 0.755 $\pm$ 0.026 \\
TGSP+SVM & 0.811 & 0.808 $\pm$ 0.009 & 0.729 $\pm$ 0.020 \\
MDM & 0.644 & 0.660 $\pm$ 0.004 & 0.719 $\pm$ 0.020 \\
EEGNet & 0.824 & 0.823 $\pm$ 0.014 & 0.505 $\pm$ 0.159 \\
\bottomrule
\end{tabular}
\end{table}

\begin{table}[H]
\centering
\caption{Within-user augmentation, per seed --- Seed 1.}
\label{tab:t21-s1}
\small
\begin{tabular}{@{}lccc@{}}
\toprule
Classifier & Baseline & with Aug & Aug only \\
\midrule
CSP+LDA & 0.769 & 0.756 $\pm$ 0.015 & 0.701 $\pm$ 0.022 \\
TGSP+SVM & 0.806 & 0.809 $\pm$ 0.008 & 0.710 $\pm$ 0.037 \\
MDM & 0.650 & 0.673 $\pm$ 0.012 & 0.708 $\pm$ 0.021 \\
EEGNet & 0.822 & 0.853 $\pm$ 0.028 & 0.404 $\pm$ 0.124 \\
\bottomrule
\end{tabular}
\end{table}

\begin{table}[H]
\centering
\caption{Within-user augmentation, per seed --- Seed 2.}
\label{tab:t21-s2}
\small
\begin{tabular}{@{}lccc@{}}
\toprule
Classifier & Baseline & with Aug & Aug only \\
\midrule
CSP+LDA & 0.789 & 0.800 $\pm$ 0.010 & 0.768 $\pm$ 0.011 \\
TGSP+SVM & 0.819 & 0.807 $\pm$ 0.012 & 0.731 $\pm$ 0.035 \\
MDM & 0.650 & 0.654 $\pm$ 0.022 & 0.653 $\pm$ 0.036 \\
EEGNet & 0.849 & 0.848 $\pm$ 0.004 & 0.536 $\pm$ 0.059 \\
\bottomrule
\end{tabular}
\end{table}

\begin{table}[H]
\centering
\caption{Within-user, individual augmentation draws underlying Table~\ref{tab:t21-s0} -- \ref{tab:t21-s2} --- Seed 0.}
\label{tab:t22-s0}
\resizebox{\textwidth}{!}{%
\begin{tabular}{@{}lccccccccc@{}}
\toprule
Classifier & Baseline & With Aug 1 & 2 & 3 & 4 & Aug Only 1 & 2 & 3 & 4 \\
\midrule
CSP+LDA & 0.769 & 0.783 & 0.772 & 0.786 & 0.797 & 0.753 & 0.778 & 0.750 & 0.739 \\
TGSP+SVM & 0.811 & 0.800 & 0.814 & 0.808 & 0.808 & 0.728 & 0.714 & 0.744 & 0.731 \\
MDM & 0.644 & 0.658 & 0.661 & 0.664 & 0.658 & 0.708 & 0.733 & 0.708 & 0.725 \\
EEGNet & 0.824 & 0.831 & 0.814 & 0.831 & 0.817 & 0.417 & 0.561 & 0.425 & 0.617 \\
\bottomrule
\end{tabular}%
}
\end{table}

\begin{table}[H]
\centering
\caption{Within-user, individual augmentation draws --- Seed 1.}
\label{tab:t22-s1}
\resizebox{\textwidth}{!}{%
\begin{tabular}{@{}lccccccccc@{}}
\toprule
Classifier & Baseline & With Aug 1 & 2 & 3 & 4 & Aug Only 1 & 2 & 3 & 4 \\
\midrule
CSP+LDA & 0.769 & 0.744 & 0.767 & 0.756 & 0.758 & 0.711 & 0.681 & 0.708 & 0.703 \\
TGSP+SVM & 0.806 & 0.806 & 0.808 & 0.817 & 0.806 & 0.722 & 0.694 & 0.686 & 0.736 \\
MDM & 0.650 & 0.669 & 0.681 & 0.664 & 0.678 & 0.722 & 0.692 & 0.714 & 0.706 \\
EEGNet & 0.822 & 0.861 & 0.831 & 0.850 & 0.872 & 0.353 & 0.450 & 0.325 & 0.489 \\
\bottomrule
\end{tabular}%
}
\end{table}

\begin{table}[H]
\centering
\caption{Within-user, individual augmentation draws --- Seed 2.}
\label{tab:t22-s2}
\resizebox{\textwidth}{!}{%
\begin{tabular}{@{}lccccccccc@{}}
\toprule
Classifier & Baseline & With Aug 1 & 2 & 3 & 4 & Aug Only 1 & 2 & 3 & 4 \\
\midrule
CSP+LDA & 0.789 & 0.797 & 0.800 & 0.808 & 0.794 & 0.775 & 0.772 & 0.764 & 0.761 \\
TGSP+SVM & 0.819 & 0.808 & 0.817 & 0.803 & 0.800 & 0.700 & 0.742 & 0.750 & 0.733 \\
MDM & 0.650 & 0.647 & 0.675 & 0.647 & 0.647 & 0.653 & 0.683 & 0.644 & 0.631 \\
EEGNet & 0.849 & 0.847 & 0.850 & 0.850 & 0.844 & 0.531 & 0.589 & 0.503 & 0.522 \\
\bottomrule
\end{tabular}%
}
\end{table}

\subsection{Cross-User (Per Held-Out Subject)}
\label{app:crossuser-persubject}

\begin{table}[H]
\centering
\caption{Cross-user per-subject augmentation results --- Held-out Subject 1.}
\label{tab:t23-s1}
\small
\begin{tabular}{@{}llccc@{}}
\toprule
Space & Classifier & Baseline & with Aug & Aug only \\
\midrule
signal & CSP+LDA & 0.693 & 0.689 $\pm$ 0.014 & 0.651 $\pm$ 0.026 \\
signal & TGSP+SVM & 0.630 & 0.644 $\pm$ 0.014 & 0.638 $\pm$ 0.028 \\
signal & MDM & 0.572 & 0.594 $\pm$ 0.008 & 0.622 $\pm$ 0.021 \\
signal & EEGNet & 0.812 & 0.812 $\pm$ 0.035 & 0.632 $\pm$ 0.149 \\
\bottomrule
\end{tabular}
\end{table}

\begin{table}[H]
\centering
\caption{Cross-user per-subject augmentation results --- Held-out Subject 2.}
\label{tab:t23-s2}
\small
\begin{tabular}{@{}llccc@{}}
\toprule
Space & Classifier & Baseline & with Aug & Aug only \\
\midrule
signal & CSP+LDA & 0.743 & 0.737 $\pm$ 0.018 & 0.665 $\pm$ 0.038 \\
signal & TGSP+SVM & 0.752 & 0.749 $\pm$ 0.006 & 0.491 $\pm$ 0.120 \\
signal & MDM & 0.623 & 0.583 $\pm$ 0.028 & 0.522 $\pm$ 0.069 \\
signal & EEGNet & 0.768 & 0.747 $\pm$ 0.102 & 0.574 $\pm$ 0.056 \\
\bottomrule
\end{tabular}
\end{table}

\begin{table}[H]
\centering
\caption{Cross-user per-subject augmentation results --- Held-out Subject 3.}
\label{tab:t23-s3}
\small
\begin{tabular}{@{}llccc@{}}
\toprule
Space & Classifier & Baseline & with Aug & Aug only \\
\midrule
signal & CSP+LDA & 0.703 & 0.727 $\pm$ 0.027 & 0.692 $\pm$ 0.008 \\
signal & TGSP+SVM & 0.738 & 0.732 $\pm$ 0.011 & 0.642 $\pm$ 0.031 \\
signal & MDM & 0.588 & 0.610 $\pm$ 0.012 & 0.630 $\pm$ 0.020 \\
signal & EEGNet & 0.639 & 0.633 $\pm$ 0.049 & 0.353 $\pm$ 0.034 \\
\bottomrule
\end{tabular}
\end{table}

\begin{table}[H]
\centering
\caption{Cross-user per-subject augmentation results --- Held-out Subject 4.}
\label{tab:t23-s4}
\small
\begin{tabular}{@{}llccc@{}}
\toprule
Space & Classifier & Baseline & with Aug & Aug only \\
\midrule
signal & CSP+LDA & 0.761 & 0.795 $\pm$ 0.016 & 0.759 $\pm$ 0.038 \\
signal & TGSP+SVM & 0.777 & 0.772 $\pm$ 0.011 & 0.700 $\pm$ 0.034 \\
signal & MDM & 0.687 & 0.719 $\pm$ 0.024 & 0.747 $\pm$ 0.016 \\
signal & EEGNet & 0.744 & 0.757 $\pm$ 0.028 & 0.501 $\pm$ 0.124 \\
\bottomrule
\end{tabular}
\end{table}

\begin{table}[H]
\centering
\caption{Cross-user, individual augmentation draws underlying Table~\ref{tab:t23-s1}'s signal-space rows --- Held-out Subject 1.}
\label{tab:t24-s1}
\resizebox{\textwidth}{!}{%
\begin{tabular}{@{}lccccccccc@{}}
\toprule
Classifier & Baseline & With Aug 1 & 2 & 3 & 4 & Aug Only 1 & 2 & 3 & 4 \\
\midrule
CSP+LDA & 0.693 & 0.693 & 0.689 & 0.676 & 0.697 & 0.670 & 0.641 & 0.635 & 0.658 \\
TGSP+SVM & 0.630 & 0.635 & 0.656 & 0.641 & 0.645 & 0.664 & 0.626 & 0.635 & 0.628 \\
MDM & 0.572 & 0.599 & 0.589 & 0.591 & 0.597 & 0.639 & 0.612 & 0.612 & 0.624 \\
EEGNet & 0.812 & 0.823 & 0.779 & 0.823 & 0.823 & 0.616 & 0.526 & 0.633 & 0.754 \\
\bottomrule
\end{tabular}%
}
\end{table}

\begin{table}[H]
\centering
\caption{Cross-user, individual augmentation draws --- Held-out Subject 2.}
\label{tab:t24-s2}
\resizebox{\textwidth}{!}{%
\begin{tabular}{@{}lccccccccc@{}}
\toprule
Classifier & Baseline & With Aug 1 & 2 & 3 & 4 & Aug Only 1 & 2 & 3 & 4 \\
\midrule
CSP+LDA & 0.743 & 0.733 & 0.724 & 0.752 & 0.738 & 0.644 & 0.692 & 0.646 & 0.678 \\
TGSP+SVM & 0.752 & 0.754 & 0.749 & 0.747 & 0.745 & 0.570 & 0.428 & 0.425 & 0.540 \\
MDM & 0.623 & 0.575 & 0.563 & 0.605 & 0.589 & 0.497 & 0.478 & 0.575 & 0.540 \\
EEGNet & 0.768 & 0.715 & 0.701 & 0.841 & 0.731 & 0.547 & 0.618 & 0.545 & 0.586 \\
\bottomrule
\end{tabular}%
}
\end{table}

\begin{table}[H]
\centering
\caption{Cross-user, individual augmentation draws --- Held-out Subject 3.}
\label{tab:t24-s3}
\resizebox{\textwidth}{!}{%
\begin{tabular}{@{}lccccccccc@{}}
\toprule
Classifier & Baseline & With Aug 1 & 2 & 3 & 4 & Aug Only 1 & 2 & 3 & 4 \\
\midrule
CSP+LDA & 0.703 & 0.703 & 0.730 & 0.741 & 0.736 & 0.698 & 0.687 & 0.692 & 0.690 \\
TGSP+SVM & 0.738 & 0.725 & 0.738 & 0.727 & 0.738 & 0.634 & 0.639 & 0.625 & 0.670 \\
MDM & 0.588 & 0.608 & 0.619 & 0.612 & 0.601 & 0.612 & 0.641 & 0.632 & 0.634 \\
EEGNet & 0.639 & 0.603 & 0.672 & 0.614 & 0.643 & 0.339 & 0.381 & 0.357 & 0.335 \\
\bottomrule
\end{tabular}%
}
\end{table}

\begin{table}[H]
\centering
\caption{Cross-user, individual augmentation draws --- Held-out Subject 4.}
\label{tab:t24-s4}
\resizebox{\textwidth}{!}{%
\begin{tabular}{@{}lccccccccc@{}}
\toprule
Classifier & Baseline & With Aug 1 & 2 & 3 & 4 & Aug Only 1 & 2 & 3 & 4 \\
\midrule
CSP+LDA & 0.761 & 0.786 & 0.809 & 0.791 & 0.793 & 0.777 & 0.772 & 0.763 & 0.724 \\
TGSP+SVM & 0.777 & 0.779 & 0.772 & 0.772 & 0.763 & 0.720 & 0.717 & 0.685 & 0.678 \\
MDM & 0.687 & 0.740 & 0.708 & 0.720 & 0.708 & 0.759 & 0.743 & 0.752 & 0.736 \\
EEGNet & 0.744 & 0.779 & 0.752 & 0.738 & 0.761 & 0.405 & 0.595 & 0.508 & 0.494 \\
\bottomrule
\end{tabular}%
}
\end{table}

\end{document}